\documentclass{article}
\usepackage[preprint]{log_2023}	
\usepackage{microtype}
\usepackage{graphicx}
\usepackage{subfigure}
\usepackage{booktabs} 
\usepackage{wrapfig}
\usepackage{enumitem}

\usepackage[utf8]{inputenc} 
\usepackage[T1]{fontenc}    

\usepackage{url}            
\usepackage{booktabs}       
\usepackage{amsfonts}       
\usepackage{nicefrac}       
\usepackage{microtype}      
\usepackage{xspace} 
\usepackage{amsmath}
\usepackage{caption}
\usepackage{wrapfig}
\usepackage{multirow}

\newcommand{\eg}{\emph{e.g.,}\xspace}
\newcommand{\ie}{\emph{i.e.,}\xspace}

\title{Do Neural Scaling Laws Exist on Graph Self-Supervised Learning?}

\usepackage[textsize=tiny]{todonotes}

\author{%
\normalsize
Qian Ma\textsuperscript{\textmd{1}}  \quad Haitao Mao\textsuperscript{\textmd{2}} \quad Jingzhe Liu \textsuperscript{\textmd{2}} \quad Zhehua Zhang \textsuperscript{\textmd{1}} \quad  \\ \textbf{Chunlin Feng\textsuperscript{\textmd{1}} \quad Yu Song\textsuperscript{\textmd{2}} \quad Yihan Shao\textsuperscript{\textmd{1}}
    \quad Yao Ma\textsuperscript{\textmd{1}}} \\
  \normalsize
  \textsuperscript{\textmd{1}} Rensselaer Polytechnic Institute \quad \textsuperscript{\textmd{2}} Michigan State University\\
  \{maq5, zhangz45, fengc5, shaoy9,  may13\}@rpi.edu,\\ \{ haitaoma, liujin33, songyu5\}@msu.edu\\
}

\begin{document}

\maketitle

\begin{abstract}
  Self-supervised learning~(SSL) is essential to obtain foundation models in NLP and CV domains via effectively leveraging knowledge in large-scale unlabeled data. The reason for its success is that a suitable SSL design can help the model to follow the neural scaling law, i.e., the performance consistently improves with increasing model and dataset sizes. However, it remains a mystery whether existing SSL in the graph domain can follow the scaling behavior toward building Graph Foundation Models~(GFMs) with large-scale pre-training. In this study, we examine whether existing graph SSL techniques can follow the neural scaling behavior with the potential to serve as the essential component for GFMs. Our benchmark includes comprehensive SSL technique implementations with analysis conducted on both the conventional SSL setting and many new settings adopted in other domains. Surprisingly, despite the SSL loss continuously decreasing, no existing graph SSL techniques follow the neural scaling behavior on the downstream performance. The model performance only merely fluctuates on different data scales and model scales. Instead of the scales, the key factors influencing the performance are the choices of model architecture and pretext task design. This paper examines existing SSL techniques for the feasibility of Graph SSL techniques in developing GFMs and opens a new direction for graph SSL design with the new evaluation prototype. Our code implementation is available online to ease reproducibility~\footnote{https://github.com/GraphSSLScaling/GraphSSLScaling}. 
\end{abstract}

\section{Introduction\label{sec:intro}}

Self-supervised learning~(SSL)~\cite{jaiswal2020survey} is to leverage the informative patterns from abundant unlabeled data via pre-training. 
SSL techniques serve an indispensable role in building Foundation Models in CV and NLP domains with successful applications~\cite{VIT,BERT,nlp_scaling}. 
A successful SSL design can observe the neural scaling law behavior where the test performance can continuously improve and the test loss continuously decreases with increasing pre-training data size and the model parameter size~\cite{alabdulmohsin2022revisiting}.
Neural scaling laws serve as the key principle for the success of foundation models in CV and NLP domains. 

SSL techniques~\cite{GraphMAE22,GraphCL20,sun2019infograph} are also successfully adopted in the graph domain while there is no Graph Foundation Model~(GFM) with SSL so far . 
It remains unclear whether graph SSL techniques follow the scaling law. To this end, we benchmark existing graph SSL techniques to examine whether they can follow the neural scaling behavior with the potential to build GFMs~\cite{mao2024graph, galkin2023towards}. 
We focus on the graph classification task instead of transductive node classification and link prediction as the unidentified relationship between train and test nodes.
An inductive graph classification setting helps to construct clear-control data scaling settings. 
Initial observations demonstrate that the graph SSL loss continuously decreases on the test set with increasing data scale and model scale. 
However, despite the decreasing SSL loss, the downstream task performance does not observe the scaling behavior correspondingly. 
Instead of data scale and model scale, key factors that influence the downstream performance are the non-parametric aggregations derived from model architecture design and the SSL objective design.
Such observations illustrate that the current graph SSL objective may not be a good choice for training a GFM~\cite{mao2024graph, galkin2023towards}. 
Therefore, we introduce a new evaluation perspective on scaling law. The setting helps state a new position on the graph SSL design in the GFM era towards better scaling.

\textbf{Organizations.} The main focus of this work is to explore the scaling law on the existing GraphSSL methods from both data scaling and model scaling. We tend to examine whether the scaling law can be applied to the existing GraphSSL methods so that they can have the potential to serve as a part of the Graph Foundation Model.
The following sections will be arranged in the following way:
In Section~\ref{sec:related_works}, we briefly introduce the neural scaling law and the SSL learning on Graph.
In Section~\ref{sec:experiments}, we introduce the basis settings of our experiments and the existing GraphSSL methods to investigate.
In Section~\ref{sec:data_scaling}, we present the results related to data scaling, and our analysis reveals that there is a gap between the SSL pre-training task and the downstream task resulting in the vanishing of data scaling law.
In Section~\ref{sec:model_scaling}, we present the results related to model scaling. Our observations indicate that the model architecture is a key factor influencing the model's performance on SSL tasks instead of the simple number of parameters or model scale.

\section{Related Works\label{sec:related_works}}

\textbf{Neural Scaling Law}
The general idea of the neural scaling law is that the model’s performance will keep improving with the scaling of training data or model parameters~\cite{kaplan2020scaling}. 
The quantitive formulation of the Neural Scaling Law is typically described in a power-law form as follows, which is first proposed by Hestness et al~\cite{hestness2017deep}.
\begin{equation}
    \mathbf{\epsilon  = aX^{-b} + \epsilon_\infty}
\end{equation}

The variable $\mathbf{X}$ represents the size of the model or the training set. 
The $\mathbf{\epsilon}$ is the prediction error of the model. $\mathbf{a}$, $\mathbf{b}$ and $\mathbf{\epsilon_\infty} > 0$ are all positive parameters. 
Under the guidance of neural scaling law, researchers could predict the performance of large models based on small-scale experiments, which greatly saves the costs of redundant runs.
Moreover, the scaling laws can be applied to benchmark different models for the backbone of foundation models.
Hence, neural scaling law has helped the development of large models in computer vision~\cite{hestness2017deep, henighan2020scaling, sharma2022scaling, Zhai_2022_CVPR} and natural language processing~\cite{kaplan2020scaling,henighan2020scaling,hoffmann2022training,gordon2021data,ghorbani2021scaling,pmlr-v162-bansal22b,fernandes2023scaling}.

Liu et al~\cite{liu2024neural} take an initial step of developing the neural scaling laws in the general graph domain. Specifically, it verifies the general forms of neural scaling laws on graphs. It also discovers some unique phenomena of model scaling and proposes a proper metric for data scaling on graphs. 
Within specific graph domains \eg molecular graphs, there are existing works~\cite{molescale} that discovered the scaling of GNNs.
These works provide a foundation for our study but are limited to supervised learning, while our focus is self-supervised learning.

\textbf{Self-Supervised Learning and its applications on Graph}.
The rise of self-supervised learning in Natural Language Processing (NLP) and Computer Vision (CV)~\cite{CLNLP,CLCV1,CLCV2} has shifted attention to learning paradigms that do not depend on annotated data. 
The burgeoning interest in self-supervised learning methodologies presents an invaluable opportunity for graph learning research, particularly in overcoming the reliance on annotated data. A growing body of work has introduced a variety of self-supervised learning strategies for graph data~\cite{DGI18,GRACE20,GCA21,CCA21,BGRL21,SUGRL22,SFA23,GraphMAE22}, marking a critical evolution in the field. These methods aim to reproduce the success of self-supervised learning in graph learning research. 
However, whether scaling law exists under these Graph SSL methods are still mysterious.

By leveraging abundant unlabeled data in the real world and expanding model scale, there are a lot of models~\cite{VIT,BERT} in CV, and NLP areas that serve as foundation models as they benefit from the scaling law during the pre-training stage.
If there exists a GraphSSL method following the neural scaling law, we believe that it has a solid basis to serve as a part of the graph foundation model.

\section{Experiment Setups\label{sec:experiments}}

To ensure the comprehensiveness of our exploration, we implement the existing representative Graph SSL methods on various datasets. We select graph classification as the downstream task for evaluation.
Here we provide some basic description of the methods and datasets, more details can be found in the Appendix~\ref{app:dataset}.

\noindent{\bf Graph SSL Methods.}
We conducted experiments on the following Graph SSL methods.
{\bf (1) InfoGraph:} As a pioneer work of Graph SSL, InfoGraph ~\cite{sun2019infograph} maximizes mutual information between global graph embeddings and local sub-structure embeddings, leveraging JSD as its contrastive loss.
{\bf (2) GraphCL:} GraphCL~\cite{GraphCL20} is a general contrastive learning framework. By maximizing the representations similarity between two different randomly perturbed local sub-graphs of the same node, the encoder can be pre-trained in a SSL manner.
{\bf (3) JOAO:} JOAO~\cite{you2021graph} can automatically and dynamically select augmentations during GraphCL training.
{\bf (4) GraphMAE:} GraphMAE~\cite{hou2022graphmae} is a generative SSL methods, which aims at reconstructing the feature and information of the data. The encoder of GraphMAE is trained by reconstructing the masked data feature with provided context.
\noindent{\bf Datasets.}
We used the following Datasets for conducting experiments.
\textbf{reddit-threads}~\cite{morris2020tudataset} contains graphs presenting the task to predict whether a thread is discussion-based.
\textbf{ogbg-molhiv},\textbf{ogbg-molpcba} are curated by ogb~\cite{hu2020ogb}, all of them are molecular property prediction datasets and the task performance metrics are ROC-AUC and AP correspondingly.
\textbf{Experiments Settings.}
In this part, we introduce some details of our experiment settings and the differences compared to the existing evaluation protocol.
We pre-train the encoder with the existing Graph SSL methods using unlabeled data.We evaluate the GraphSSL methods by applying the pre-trained encoder to downstream task via linear probing.
\noindent \textbf{Data Split.} 
All datasets are split with the ratio 8:1:1 for training, validation, and testing set and only the pre-split training set is used to pre-train the model. To ensure the reproducibility of our experiments, we fix the split for all datasets and all methods will use the same split for experiments. 
\noindent \textbf{Evaluation Protocols.} 
In this work, we only use the split training set for pre-training to ensure the testing set used for evaluation will not be leaked in the pre-training stage.
For the evaluation of SSL methods, we follow the existing setting in our main experiments by fixing the pre-trained encoder  
and use the embeddings obtained by this specific fixed pre-trained encoder to conduct a downstream task, such as training a classifier for graph classification.
The pre-training data and downstream data are from the identical dataset.
Then the metrics on the downstream task will be reported to serve as the performance of the pre-trained encoder to reflect the feasibility of the SSL method correspondingly. 
Specifically, for the downstream task settings, we only used the pre-first 10\% of the split training data with labels.
We stored the model pre-trained after 100 epochs to further evaluate them with the downstream tasks. We also include more details in the Appendix~\ref{app:settings}.

\section{Data scaling}\label{sec:data_scaling}
In this section, we conduct experiments to explore the data scaling of Graph SSL methods and our findings.
The main observation is that there is no obvious downstream performance gain along with the scaling-up data, indicating no data-scaling effect.

\subsection{Data Scaling on Downstream Performance}\label{sec:data_scaling_perf}
To explore how the performance of downstream tasks improves with the scale-up pre-training data for Graph SSL methods, we introduce the settings and then present our observations.

{\bf Settings.} 
To verify the existence of the data scaling phenomenon of GraphSSL methods, we construct the following pipeline for pre-training and evaluation to verify data scaling. 
\begin{itemize}[leftmargin=*]
    \item For a reasonable data scaling setting, we gradually increase the ratio for pre-training data with a fixed interval by containing all data used in the previous ratios. 
    \item For each dataset, we further slice the pre-split training data with the fixed interval with 0.1 as different pre-training data ratio settings. The order of indices is fixed after generation, so we can gradually increase the data ratio for pre-training from 0.1 to 1 by slicing the indices to make sure the data from previous lower ratio can be included.
    \item For the evaluation on the downstream tasks, we trained an SVM classifier with the pre-trained model fixed following the existing protocols and reported its performance on the held-out test set as the metrics for evaluation.
    \item To examine whether the Graph SSL methods can consistently exhibit the scaling effect, we fit the equation of scaling law to our empirical results on different data scales with the least square and calculate the coefficient of determination $R^2$ for examining the quality of the fitting to the scaling.
\end{itemize}

\textbf{Observation 1. With gradually scaled-up pre-training data, no obvious scaling effect can be observed from the downstream performance.}

\begin{figure}[htbp]
    \centering
    \subfigure[\small GraphCL $R^2=0.0$\label{fig:data_GraphCL_ogbg-molhiv_perf}]{
        \begin{minipage}[b]{0.23\textwidth}
            \includegraphics[width=\columnwidth]{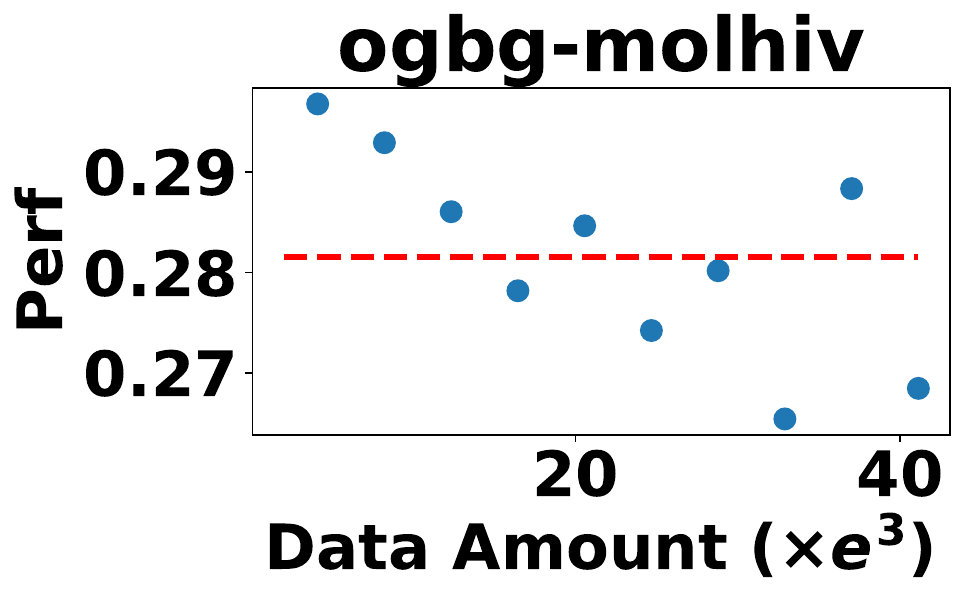}
        \end{minipage}
    }
    \subfigure[\small GraphMAE $R^2=0.31$\label{fig:data_GraphMAE_ogbg-molhiv_perf}]{
        \begin{minipage}[b]{0.23\textwidth}
            \includegraphics[width=\columnwidth]{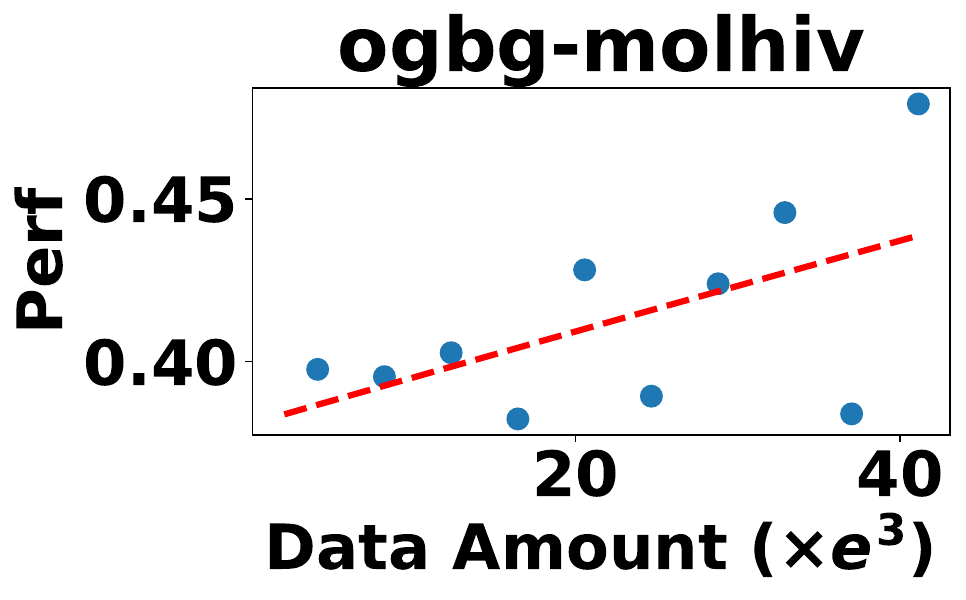}
        \end{minipage}
    }
    \subfigure[InfoGraph $R^2=0.0$\label{fig:data_InfoGraph_ogbg-molhiv_perf}]{
        \begin{minipage}[b]{0.23\textwidth}
            \includegraphics[width=\columnwidth]{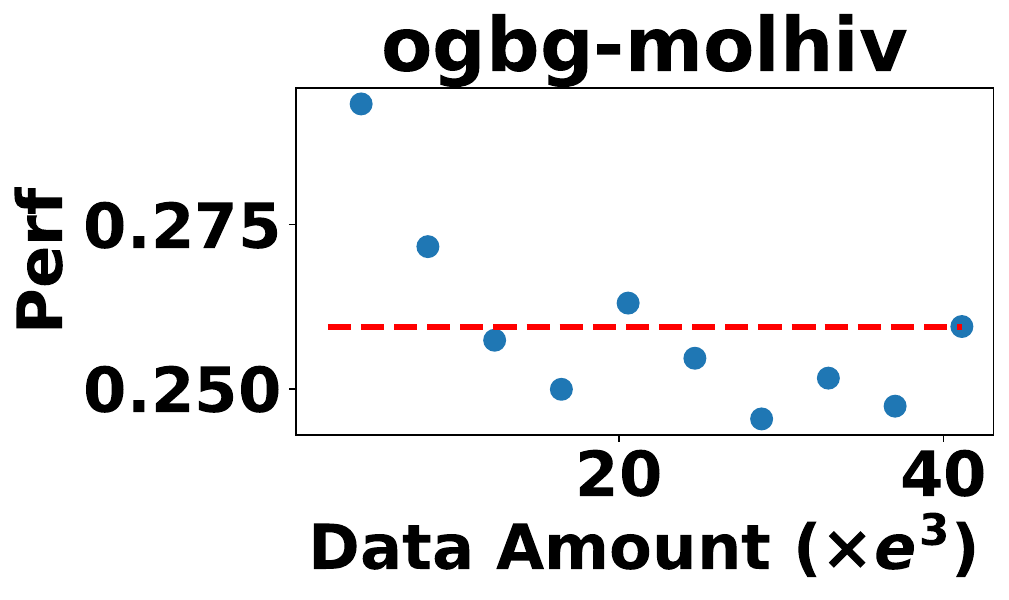}
        \end{minipage}
    }
    \subfigure[JOAO $R^2=0.0$\label{fig:data_JOAO_ogbg-molhiv_perf}]{
        \begin{minipage}[b]{0.23\textwidth}
            \includegraphics[width=\columnwidth]{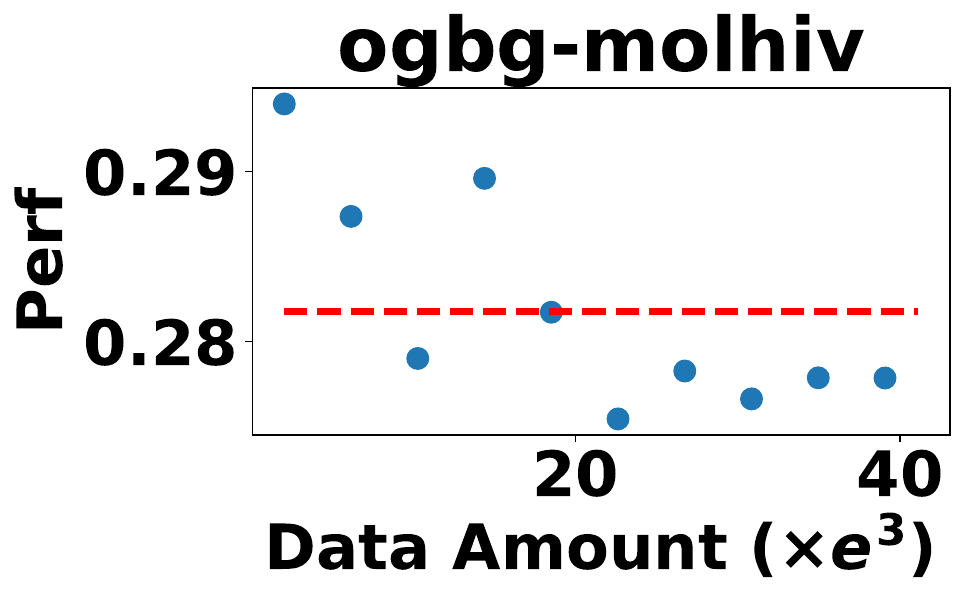}
        \end{minipage}
    }
    \caption{{Data Scaling of Performance on ogbg-molhiv.x-axis indicates the data amount used for pre-training and y-axis indicates the downstream performance. No obvious scaling behavior can be observed.}}
    \label{fig:Data_Perf_hiv}
\end{figure}

\begin{figure}[htbp]
    \centering
    \subfigure[GraphCL $R^2=0.0$]{
        \includegraphics[width=0.23\textwidth]{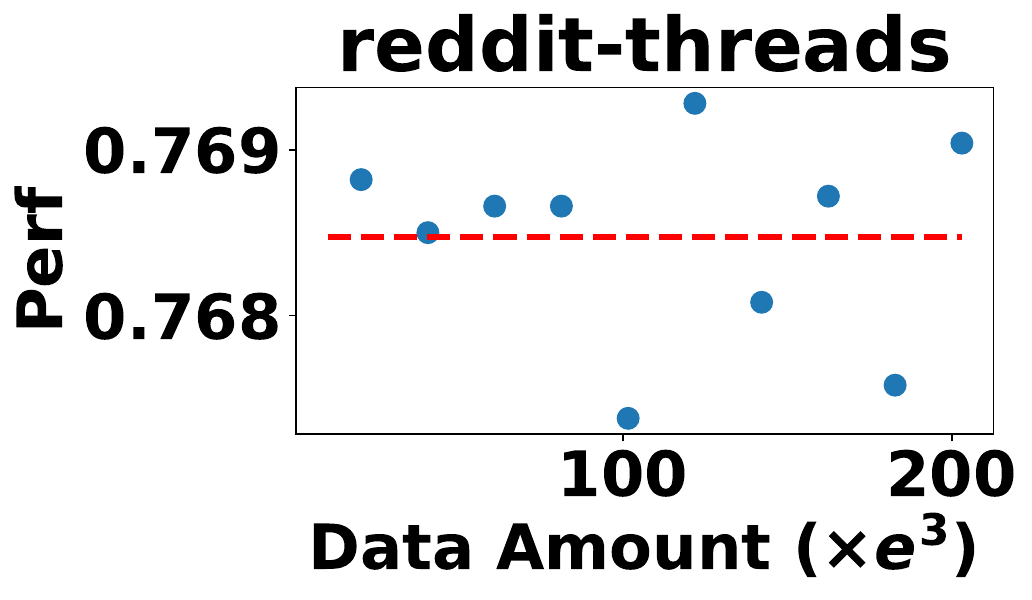} 
        \label{fig:data_GraphCL_reddit_threads_perf}
    }
    \subfigure[GraphMAE $R^2=0.0$]{
        \includegraphics[width=0.23\textwidth]{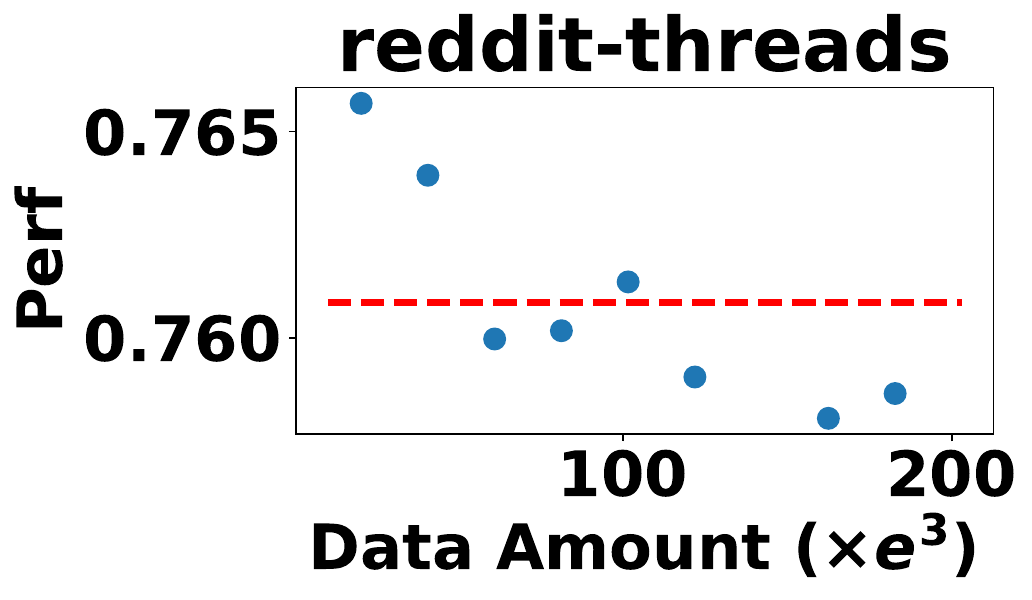}
        \label{fig:data_GraphMAE_reddit_threads_perf}
    }
    \subfigure[InfoGraph $R^2=0.0$]{
        \includegraphics[width=0.23\textwidth]{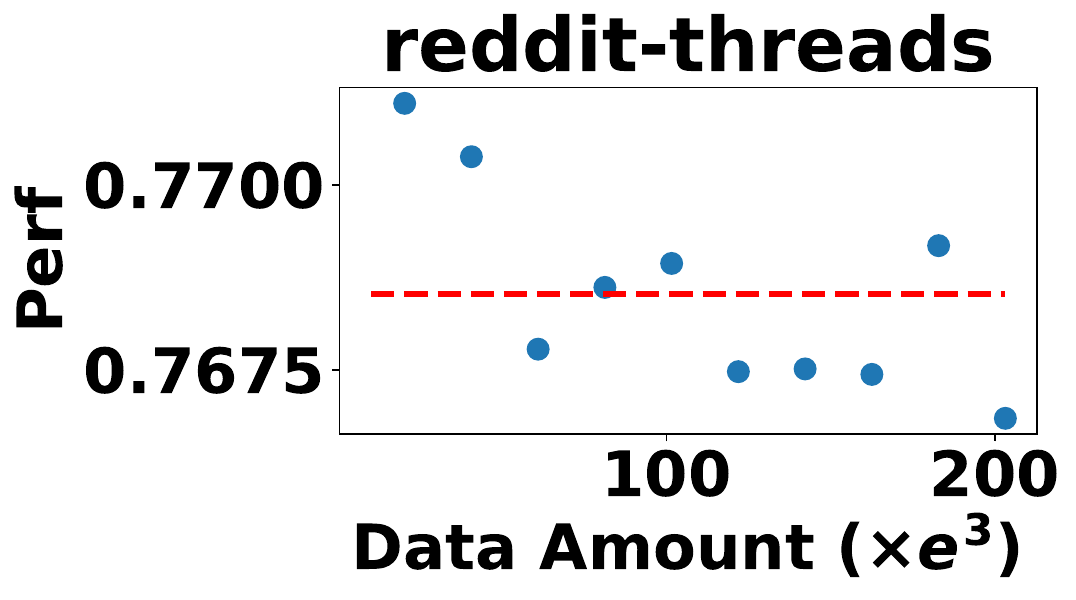} 
        \label{fig:data_InfoGraph_reddit_threads_perf}
    }
    \subfigure[JOAO $R^2=0.009$]{
        \includegraphics[width=0.23\textwidth]{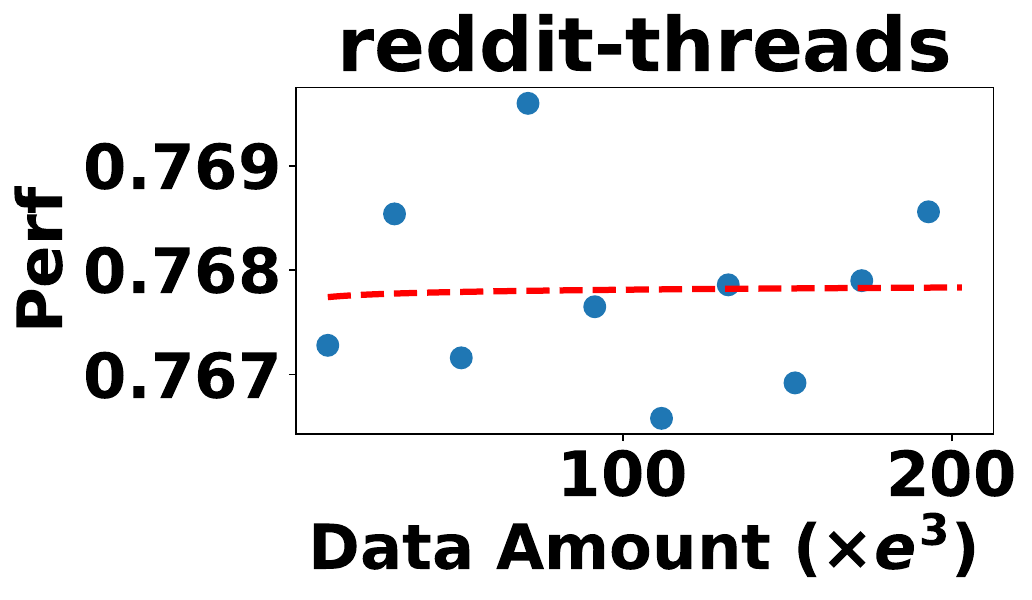}
        \label{fig:data_JOAO_oreddit_threads_perf}
    }
    \caption{Data Scaling of Performance on reddit-threads. x-axis indicates the data amount used for pre-training and y-axis indicates the downstream performance. No obvious scaling behavior can be observed.}
    \label{fig:Data_Perf_reddit}
\end{figure}

We conduct experiments to gradually increase the data ratio for pre-training to observe whether the performance gain, along with the increasing data, serves as evidence for the data-scaling behavior.  
Figure~\ref{fig:Data_Perf_hiv} and~\ref{fig:Data_Perf_reddit} illustrate the performance on reddit-threads and ogbg-molhiv datasets and more results on ogbg-molpcba can be found in Appendix~\ref{app:data_scaling}. The x-axis indicates the data amount for pre-training, and the y-axis indicates the downstream performance, respectively.
The data points are used to fit the parameters of the scaling law and $R^2$ is calculated to examine the overall quality of fitting.
Typically, a $R^2$ value larger than 0.5 can be considered significant.

Our key observation is that the performance of all investigated Graph SSL methods does not exhibit a consistent and obvious scaling behavior despite the consistently increased data ratio for pre-training.
There is no fitted curve nor large $R^2$ value to indicate that the downstream performance can scale up along with more pre-training data amount in Figure~\ref{fig:Data_Perf_hiv} and~\ref{fig:Data_Perf_reddit}.

The key conclusion derived from the above results is that, unlike the SSL methods in the NLP~\cite{kaplan2020scaling,henighan2020scaling,hoffmann2022training,gordon2021data,ghorbani2021scaling,pmlr-v162-bansal22b,fernandes2023scaling} and CV~\cite{hestness2017deep, henighan2020scaling, sharma2022scaling, Zhai_2022_CVPR} domains , \textit{Graph SSL methods do not observe data scaling behavior across graphs.}
A further investigation to understand why such a phenomenon happens can be found as follows.

During the pre-training stage, the Graph SSL methods will try to optimize their own SSL task objectives or SSL Losses as we introduced in Section~\ref{sec:related_works}. 
These objectives or losses are improved during the pre-training stage.
Therefore, we would like to investigate whether they benefit the downstream loss \ie whether the knowledge can be generalized to the unseen datasets.
If such a reduction can be observed on the downstream loss, then it indicates that the GraphSSL methods fail to scale up on downstream performance due to the huge gap between pre-training and downstream tasks.
Therefore, our further investigation aims to examine if the capability obtained along with increasing large-scale pre-training data doesn't \textit{correspond} to the {\bf downstream performance gain}.

\subsection{Data Scaling on SSL loss}

{\bf Settings.} We investigate the above question by changing the metrics we observed from the downstream performance to the same SSL task objectives. More specifically, we utilize the held-out test set to compute the same SSL loss used in the pre-training stage with the pre-trained model fixed.
In this way, we can examine if there is a gain on the same SSL task from improved capability obtained by scale-up pre-training data. 

\textbf{Observation 2. With gradually scaled-up pre-training data, consistent scaling behavior can be observed on the SSL loss.}

\begin{figure}[htbp]
    \centering
    \subfigure[InfoGraph $R^2$=0.92]{
            \includegraphics[width=0.23\columnwidth]{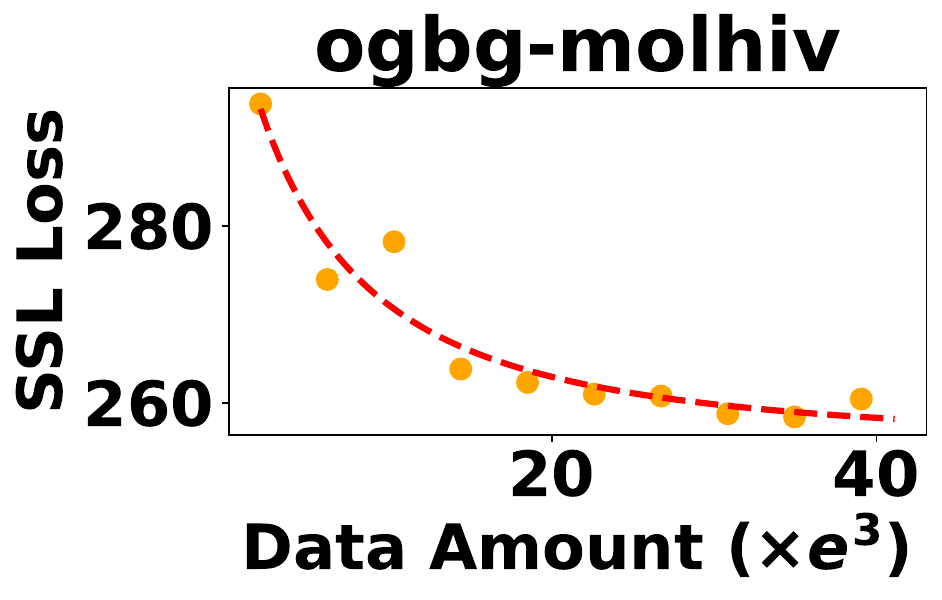} 
            \label{fig:data_InfoGraph_ogbg-molhiv_loss}
    }
    \subfigure[GraphCL $R^2$=0.51]{
        \includegraphics[width=0.23\columnwidth]{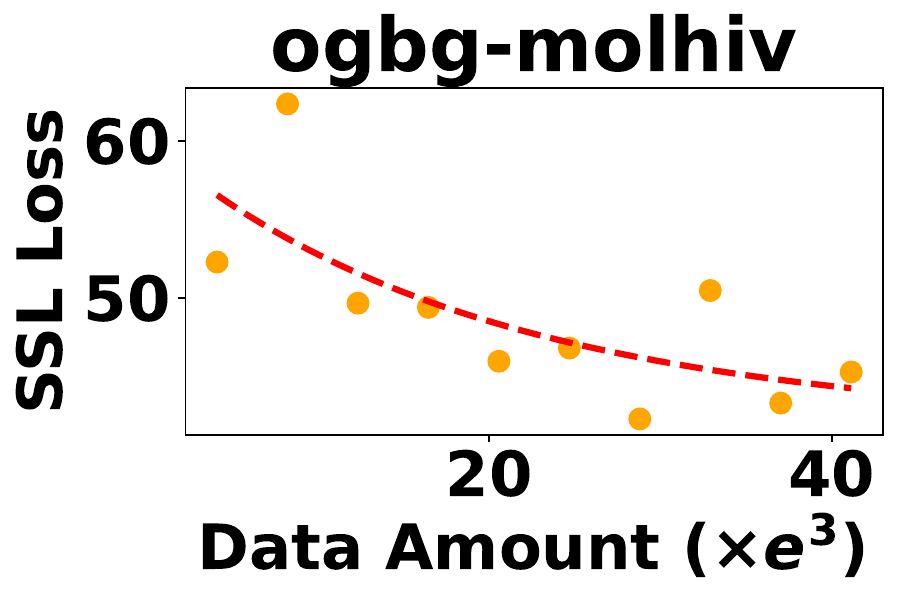} 
        \label{fig:data_GraphCL_ogbg-molhiv_loss}
    }
    \subfigure[JOAO $R^2$=0.66]{
        \includegraphics[width=0.23\textwidth]{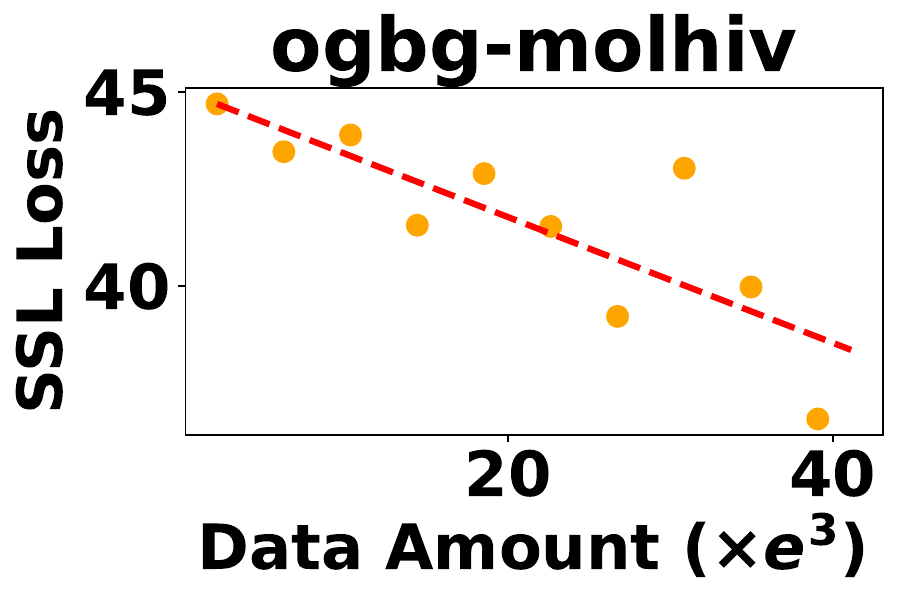} 
        \label{fig:data_JOAO_ogbg-molhiv_loss}
    }
    \subfigure[GraphMAE $R^2$=0.57]{
        \includegraphics[width=0.23\textwidth]{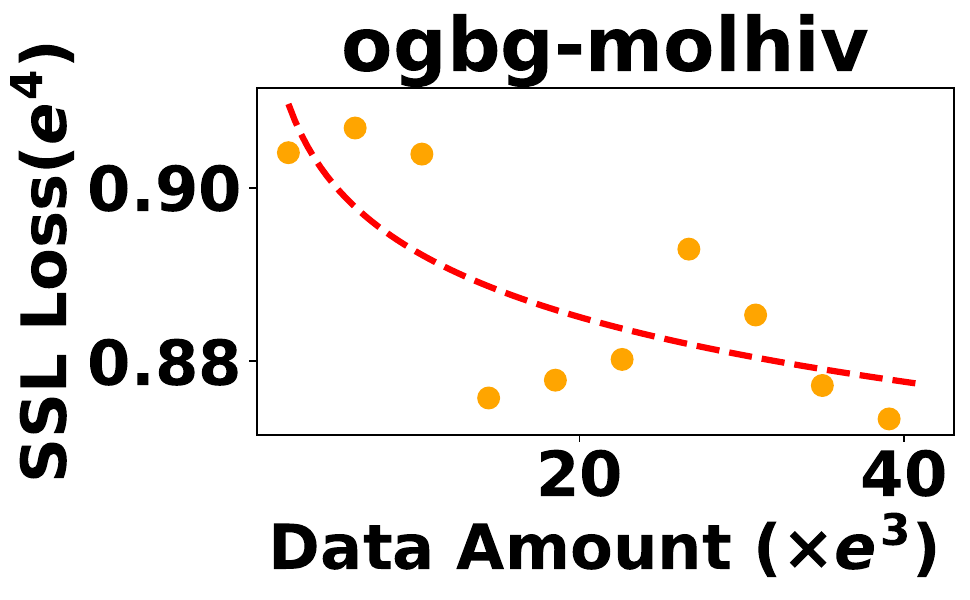} 
        \label{fig:data_GraphMAE_ogbg-molhiv_loss}
    }
    \caption{Data Scaling of SSL Loss on ogbg-molhiv.
    x-axis indicates the data amount used for pre-training and y-axis indicates the SSL Loss on the held-out test data. More obvious scaling behavior can be observed compared to performance.}
    \label{fig:data-loss-ogbg-molhiv}
\end{figure}

\begin{figure}[htbp]
    \centering
    \subfigure[InfoGraph $R^2=0.98$]{
        \includegraphics[width=0.23\textwidth]{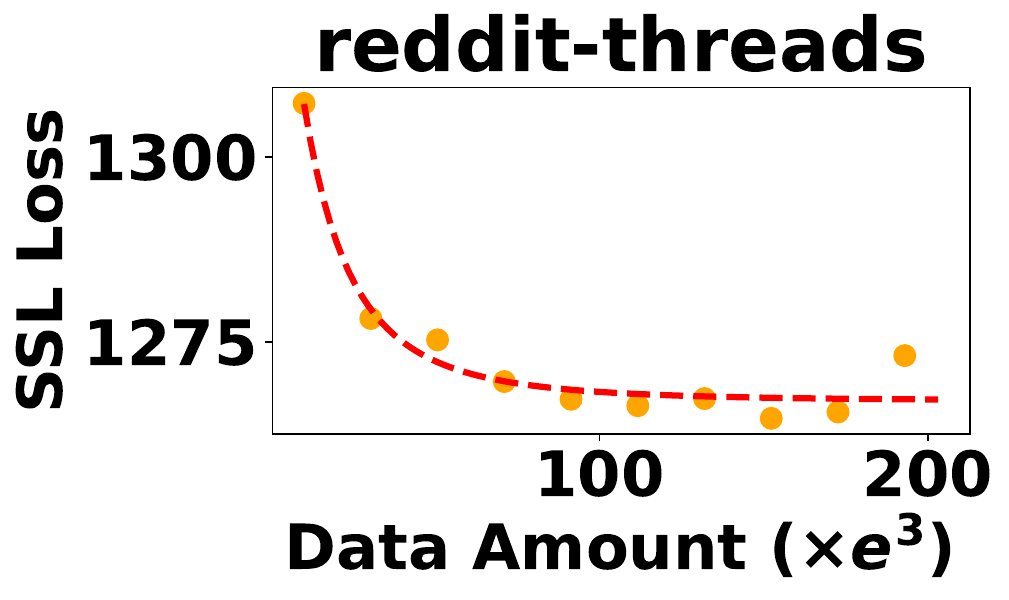}
        \label{fig:data_InfoGraph_reddit_threads_loss}
    }
    \subfigure[GraphCL $R^2=0.23$]{
        \includegraphics[width=0.23\textwidth]{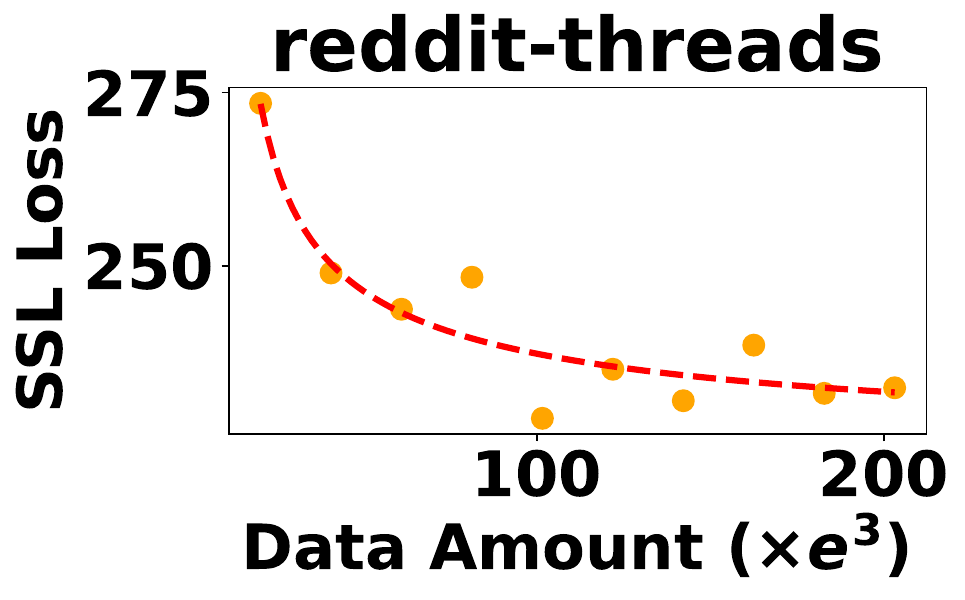}
        \label{fig:data_GraphCL_reddit_threads_loss}
    }
    \subfigure[JOAO $R^2=0.63$]{
        \includegraphics[width=0.23\textwidth]{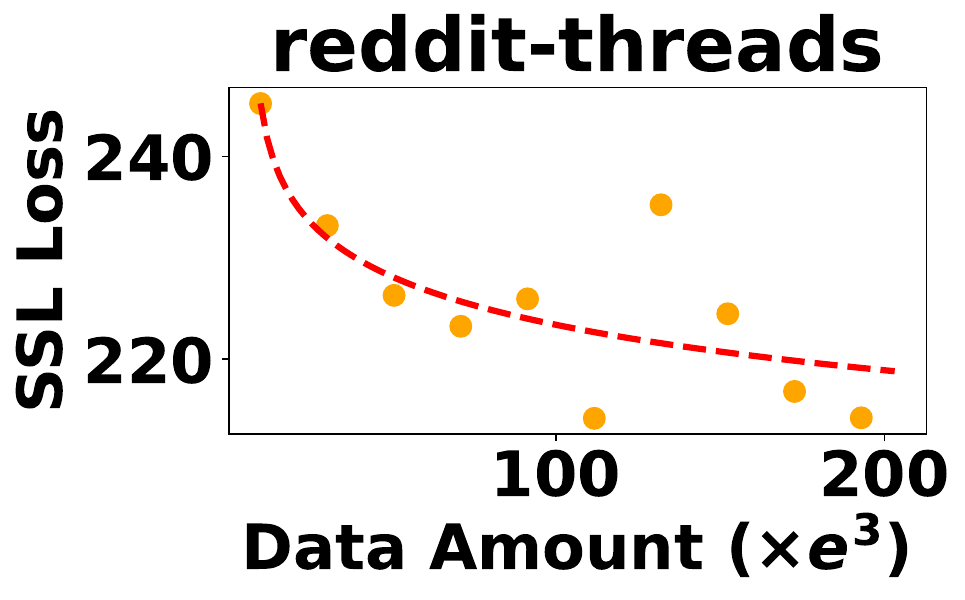}
        \label{fig:data_JOAO_reddit_threads_loss}
    }
    \subfigure[GraphMAE $R^2=0.87$]{
        \includegraphics[width=0.23\textwidth]{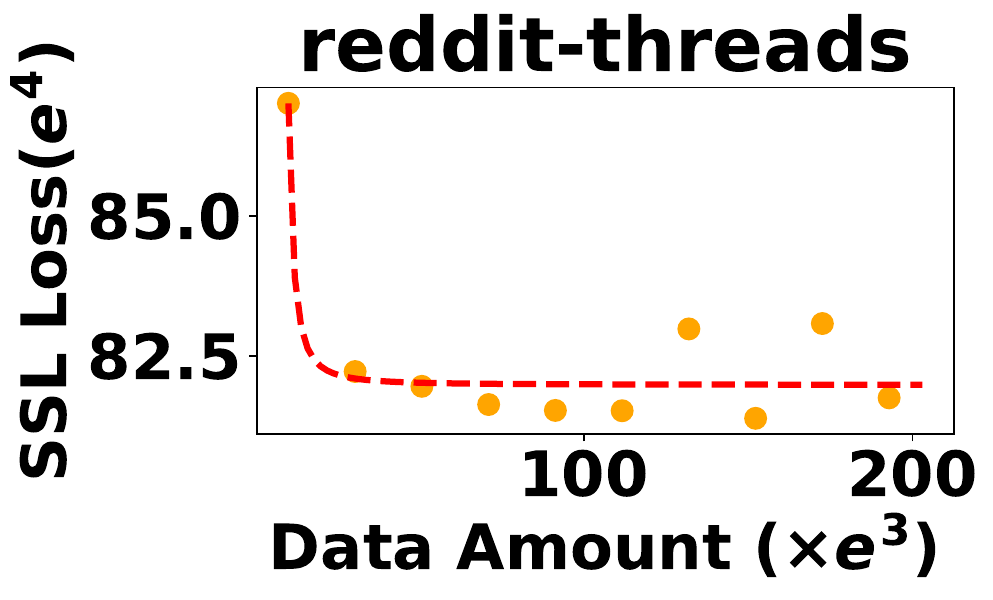}
        \label{fig:data_GraphMAE_reddit_threads_loss}
    }
    \caption{Data Scaling of SSL Loss on reddit-threads.x-axis indicates the data amount used for pre-training and y-axis indicates the SSL Loss on the held-out test data. More obvious scaling behavior can be observed compared to performance.}
    \label{fig:data-loss-reddit-threads}
\end{figure}

We conduct data scaling experiments to examine if the gain can be observed on SSL tasks with the scale-up pre-training data.
The results presented in Figure~\ref{fig:data-loss-ogbg-molhiv} and~\ref{fig:data-loss-reddit-threads} illustrate how the SSL Loss improves on reddit-threads and ogbg-molhiv datasets with scale-up pre-training data. 
The x-axis is the pre-training data amount and the y-axis is the SSL Loss. The overall fitting quality of the fitted curve obtained with the data points is examined by the $R^2$ value. 

Compared with the observation on the downstream task performance, the scaling behavior on the SSL loss is consistent and obvious. 
However, the scaling behavior could be method-specific \ie some methods behave more consistently and stably in a scaling manner while others do not.
Taking the InfoGraph as an example, as shown in Figure~\ref{fig:data_InfoGraph_ogbg-molhiv_loss}and~\ref{fig:data_InfoGraph_reddit_threads_loss} , the SSL loss evaluated on the testing data decreases as pre-training data scales.
Meanwhile, for other methods \eg GraphCL the scaling effect is less obvious and consistent as shown in Figure~\ref{fig:data_GraphCL_ogbg-molhiv_loss}and~\ref{fig:data_GraphCL_reddit_threads_loss}.

According to the above results, we observe that the scale-up pre-training data can improve the capability of SSL tasks in a data-scaling manner. 
Notably, we do not observe the scaling behavior on the downstream performance in Section~\ref{sec:data_scaling_perf}. 
Therefore, it could be the gap between the pre-training and downstream tasks that block the GraphSSL methods from following the scaling law on the downstream performance.

\section{Model Scaling}\label{sec:model_scaling}
In this section, we conduct experiments to explore the model scaling of Graph SSL methods. 
Specifically, we aim to observe how the performance improves as we increase the number of model parameters. 
Our key observation is that no consistent scaling behaviour can be observed with model scaling on performance. 

\textbf{Settings.} 
To observe the model scaling effects on the Graph SSL methods, we scale up the model parameters by increasing \textbf{(1) size of hidden dimensions} and \textbf{(2) number of layers} for the encoders applied in the Graph SSL methods.
Following the settings in the Section~\ref{sec:data_scaling}, we monitor both the graph classification downstream performance and the value of SSL objectives on the held-out test set. 

\subsection{Model Scaling on Downstream Performance}

\textbf{Observation 3. Under the two different manners of model scaling settings, there is no obvious scaling effect can be observed from the downstream performance.}

\begin{figure}
    \centering
    \includegraphics[width=\textwidth]{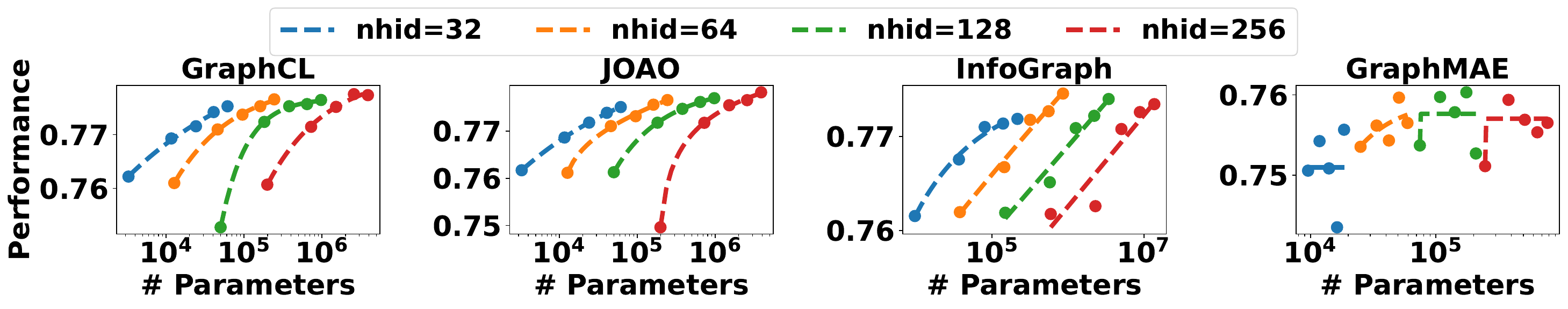}
    \caption{Performance on reddit-threads. x-axis denotes the total number of parameters and y-axis denotes the downstream performance. Obvious scaling behavior can be observed on all methods except GraphMAE. The $R^2$ values for each method are listed as follows, GraphCL:0.99,JOAO:0.99,InfoGraph:0.95,GraphMAE:0.36}
    \label{fig:model_reddit_perf}
\end{figure}

\begin{figure}
    \centering
    \includegraphics[width=\textwidth]{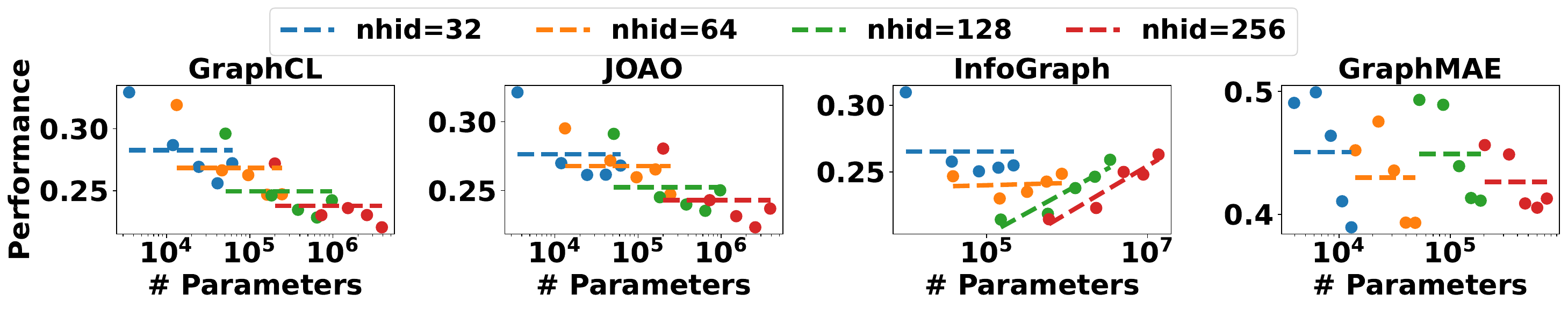}
    \caption{Performance on ogbg-molhiv. x-axis denotes the total number of parameters and y-axis denotes the downstream performance. The $R^2$ values for each method are listed as follows. GraphCL:0.0,JOAO:0.0,InfoGraph:0.42,GraphMAE=0.0}
    \label{fig:model_hiv_perf}
\end{figure}

\begin{figure}[!htbp]
    \centering
    \includegraphics[width=\textwidth]{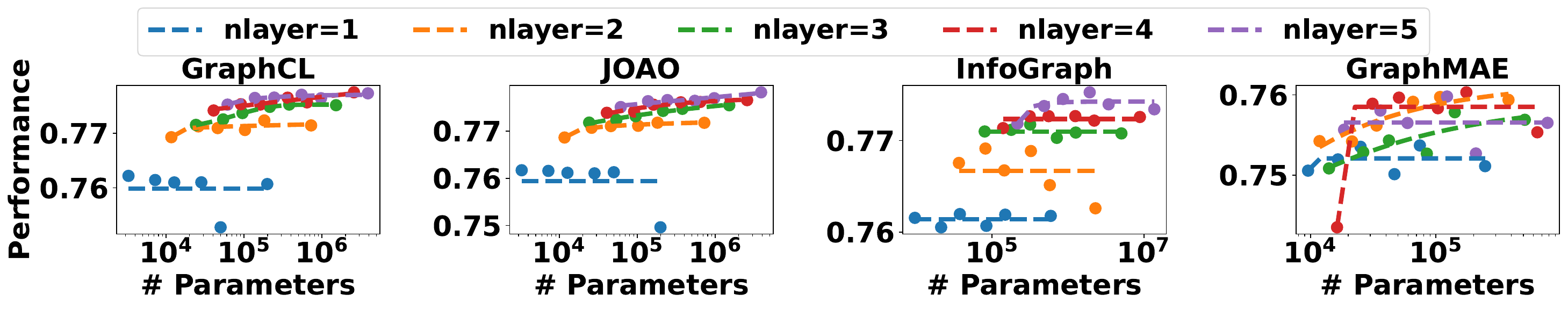}
    \caption{Performance on reddit-threads grouped by layers. x-axis denotes the total number of parameters and y-axis denotes the downstream performance. The $R^2$ values for each method are listed as follows.GraphCL:0.56,JOAO:0.79,InfoGraph:0.49,GraphMAE:0.64}
    \label{fig:model_reddit_perf_layer}
\end{figure}

\begin{figure}[!htbp]
    \centering
    \includegraphics[width=\textwidth]{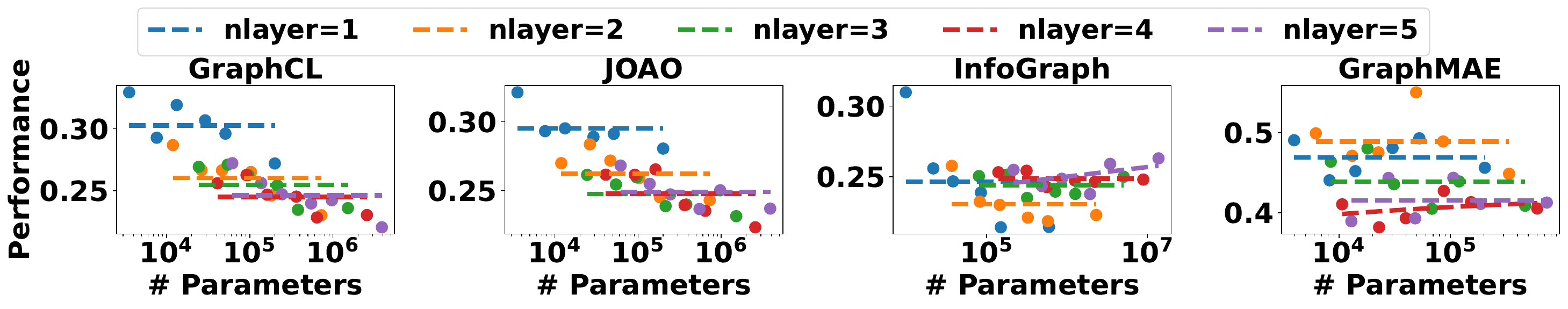}
    \caption{Performance on ogbg-molhiv grouped by layer. x-axis denotes the total number of parameters and y-axis denotes the downstream performance. No obvious scaling behaviour can be observed. The $R^2$ values for each method are listed as follows.GraphCL:0.0,JOAO:0.0,InfoGraph:0.11,GraphMAE:0.18 }
    \label{fig:model_hiv_perf_layer}
\end{figure}

We present the results showing how the downstream performance of investigated methods improve with scale-up model size on three datasets in Figure~\ref{fig:model_reddit_perf},~\ref{fig:model_hiv_perf},~\ref{fig:model_pcba_perf} and Figure~\ref{fig:model_reddit_perf_layer},~\ref{fig:model_hiv_perf_layer},~\ref{fig:model_pcba_perf_layer}  where the number of parameters is indicated by the x-axis and the downstream performance is indicated by the y-axis, and different colors indicate different settings of hidden size or number of layers. 

By increasing the number of layers with the hidden size fixed, no consistent scaling behavior can be observed as shown in Figure~\ref{fig:model_reddit_perf},~\ref{fig:model_hiv_perf},~\ref{fig:model_pcba_perf}. 
Even for GraphCL, JOAO, and InfoGraph, there seems to be an obvious scaling effect exhibited only on Reddit Threads datasets as shown in Figure~\ref{fig:model_hiv_perf}, however, this could be limited to the scale of plotting as the differences between the downstream performance metrics are very marginal and far away from being called `scaling', especially compared with the number of parameters increasing in a exponential way.
Similarly, by increasing the number of hidden size with the number of layers fixed, no scaling effect can be consistently presented for all methods across all datasets, as shown in Figure~\ref{fig:model_reddit_perf_layer},~\ref{fig:model_hiv_perf_layer},~\ref{fig:model_pcba_perf_layer}. 

Therefore, our key observation is that there is no consistent scaling behavior in the downstream performance of GraphSSL methods with either scale-up hidden dim or number of layers,
unlike the scaling effect that universally exists in CV and NLP domains by increasing the total number of parameters of the model.
Consequently, we conducted further investigation on  SSL loss to examine if the capability of scale-up model parameters benefits the SSL tasks without corresponding to the downstream performance gain. 

\subsection{Model Scaling on SSL loss}
To examine if the scaled-up model parameters can improve the capability of SSL task to reveal scaling law, we target the SSL loss on the downstream data as a metric.
{To better examine the GraphCL and JOAO, we select proper settings for data augmentation for contrastive learning as it is indicated as the key component in their original papers~\cite{GraphCL20,you2021graph}. The details are deferred to the Appendix~\ref{app:model_scaling}}

\textbf{Observation 4. Under the model scaling setting with increasing numbers of layers, the scaling effect on the SSL loss can be observed on particular datasets and SSL methods. }

\begin{figure}[!htbp]
    \centering
    \includegraphics[width=0.95\textwidth]{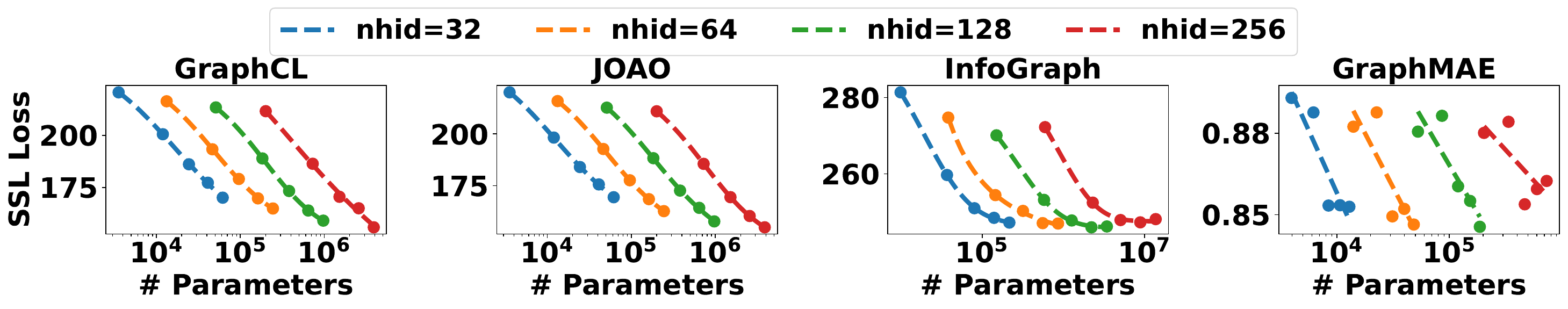}
    \caption{SSL Loss on ogbg-molhiv.Obvious scaling behaviour can be observed on all methods. x-axis denotes the total number of parameters and y-axis denotes SSL Loss on the held-out test set. The $R^2$ values for each method are listed as follows. GraphCL:0.99, JOAO:0.99, InfoGraph:0.99, GraphMAE:0.73}
    \label{fig:model_hiv_loss}
\end{figure}

\begin{figure}[!htbp]
    \centering
    \includegraphics[width=0.95\textwidth]{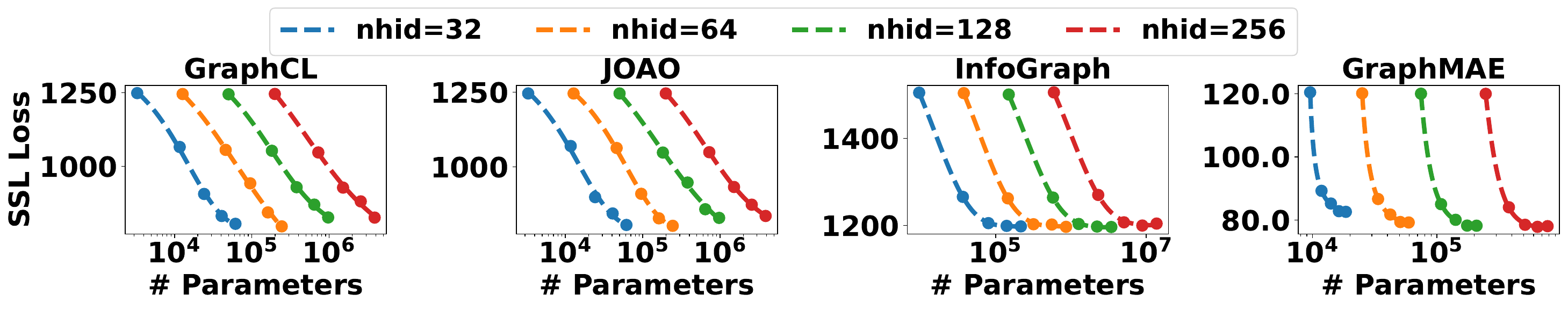}
    \caption{SSL Loss on reddit-threads. Obvious scaling behaviour can be observed on all methods. x-axis denotes the total number of parameters and y-axis denotes SSL Loss on the held-out test set. The $R^2$ values for each method are listed as follows. GraphCL:0.99, JOAO:0.99, InfoGraph:0.99, GraphMAE:0.99}
    \label{fig:model_reddit_loss}
\end{figure}

We present the results of all methods on the same dataset in Figure~\ref{fig:model_hiv_loss},~\ref{fig:model_reddit_loss} and~\ref{fig:model_pcba_loss}, where the x-axis denotes the total number of parameters of the model and the y-axis denotes the metrics of SSL loss respectively. Different colors represent different hidden size settings.

Our key conclusion from the above results is that there is method-specific model scaling behavior can be observed with the scale-up number of layers and fixed hidden size.
For \textbf{InfoGraph}, as shown in Figure~\ref{fig:model_hiv_loss}, it can exhibit more obvious scaling behavior compared to the trend observed with the performance metrics.
Moreover, the scaling behavior is consistently obvious on all other datasets as shown in Figure~\ref{fig:model_reddit_loss} and~\ref{fig:model_pcba_loss}. 
Compared with the representative contrastive SSL method InfoGraph, \textbf{GraphMAE} is a generative method, the scaling effect is not consistent nor obvious on its SSL loss \ie its feature reconstruction cosine loss. 
As shown in Figure~\ref{fig:model_hiv_loss} and~\ref{fig:model_pcba_loss}, the scaling effect indicated by the $R^2$ value is not as obvious and consistent as that presented in Figure~\ref{fig:model_reddit_loss}.
These differences indicate that the SSL task design of GraphMAE can not consistently benefit from the scaling up of model parameters. 
As contrastive SSL methods, \textbf{GraphCL} and \textbf{JOAO} can exhibit similar scaling behavior as InfoGraph on all datasets as shown in Figure~\ref{fig:model_hiv_loss},~\ref{fig:model_reddit_loss} and~\ref{fig:model_pcba_loss}.
\textbf{Observation 5. Under the model scaling settings with increasing hidden size, there is no obvious scaling effect can be observed from the SSL loss across datasets.}

\begin{figure}[!htbp]
    \centering
    \includegraphics[width=0.95\textwidth]{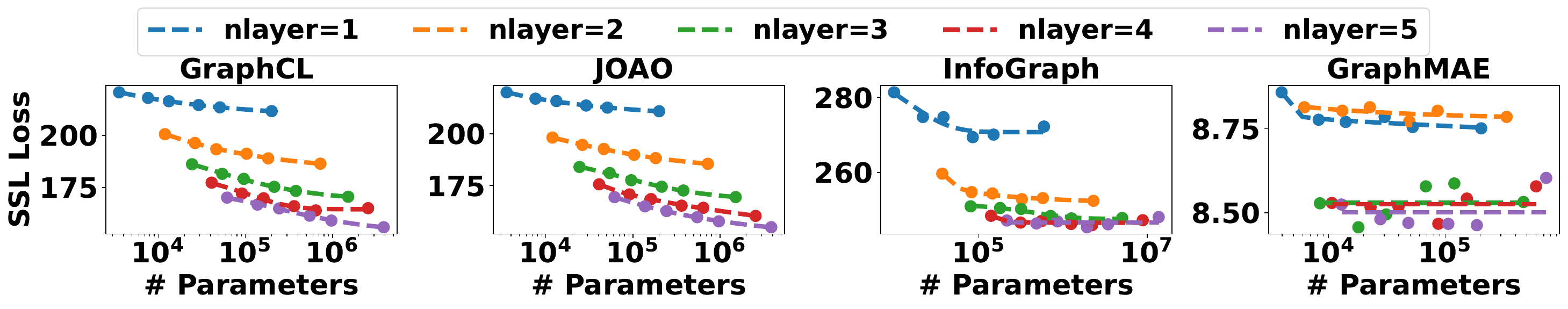}
    \caption{ SSL Loss on ogbg-molhiv grouped by layer. No obvious scaling behavior can be observed.  x-axis denotes the total number of parameters and y-axis denotes SSL Loss on the held-out test set. The $R^2$ values for each method are listed as follows. GraphCL:0.99, JOAO:0.99, InfoGraph:0.70, GraphMAE:0.26}
    \label{fig:model_hiv_loss_layer}
\end{figure}

\begin{figure}[!htbp]
    \centering
    \includegraphics[width=0.95\textwidth]{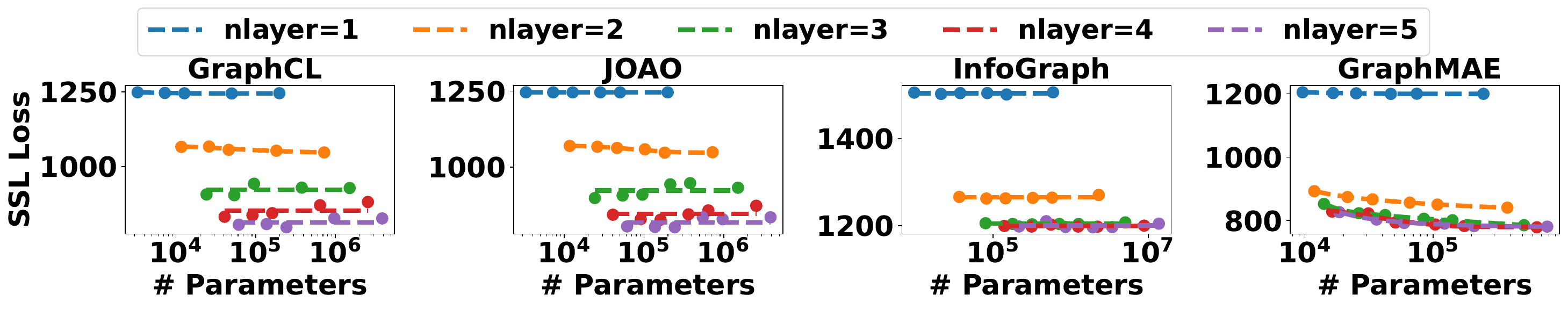}
    \caption{SSL Loss on reddit-threads grouped by layer.No obvious scaling behavior can be observed. x-axis denotes the total number of parameters and y-axis denotes SSL Loss on the held-out test set. The $R^2$ values for each method are listed as follows. GraphCL:0.36, JOAO:0.18, InfoGraph:0.00, GraphMAE:0.99. Kindly note that $R^2$ is the metric to evaluate how well the fitted curve is instead of a direct metric to indicate the existence of the scaling effect.}
    \label{fig:model_reddit_loss_layer}
\end{figure}

We also grouped the results by the same number of layers. The results are presented in Figure~\ref{fig:model_hiv_loss_layer},~\ref{fig:model_reddit_loss_layer} and~\ref{fig:model_pcba_loss_layer}, where different colors indicate different number of layers settings. 
The fitted curve and $R^2$ value to examine the fitting quality, all indicate that no consistent and obvious scaling behavior can be observed.

Our key conclusion from the above results is that there is no consistent or obvious scaling behavior can be observed with the scale-up hidden size while fixing the number of layers.
Taking InfoGraph as an example, the SSL Loss results grouped by the same number of layers shown in Figure~\ref{fig:model_reddit_loss_layer} indicates that there is no improvement on the objectives by increasing the hidden size with the number of layers fixed.
Moreover, compared to the scaling behavior exhibited by InfoGraph with scale-up number of layers in the model shown in Figure~\ref{fig:model_reddit_loss},
the different behaviors with different model scaling manners indicate that the improvement is more likely to be the result of more aggregations by stacking more layers instead of the capability of transformation with more learnable model parameters.
Consequently, we conducted a further investigation on InfoGraph for this phenomenon to examine whether the aggregation benefits the capability of InfoGraph on SSL objectives rather than the transformation with more learnable parameters.

\textbf{Observation 6. By fixing the hidden size for transformation and increasing the number of aggregations, an obvious and consistent scaling behavior can be observed on SSL Loss for the new implementation with transformation and aggregation decoupled.}

\begin{figure}
    \centering
    \includegraphics[width=\textwidth]{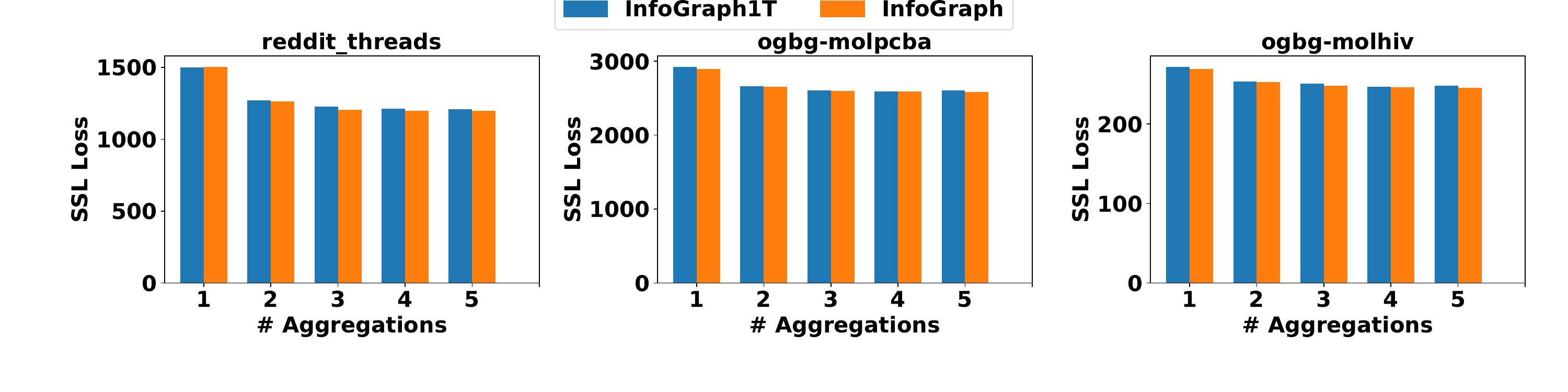}
    \caption{SSL loss Comparison on three datasets with nhid=96. x-axis denotes the number of aggregations and y-axis denotes SSL Loss on the held-out test set. Decoupled new implementation of InfoGraph with only one transformation(InfoGraph1T) can also exhibit consistent scaling behavior with marginal difference to original results}
    \label{fig:Decouple_Loss_96}
    \vspace{-2mm}
\end{figure}

To decouple the aggregation and transformation of the model, we modify the implementation of the original InfoGraph accordingly.
More specifically, we replaced the layers except for the last layers from the GINConv layer to Message Passing layers.
For the original implementation, all layers are the same, \ie for each layer, the GINConv layer will aggregate the features and then transform them.
Our new implementation only remains the one transformation layer, which is the same as the original version for feature transformation while the other layers are only message passing layers for aggregation.
In this way, the aggregation and transformation are decoupled.
The embedding obtained with the trained encoder will be fed into the projection head to map the embeddings into another latent space for calculating the SSL Loss.
Our new implementation with aggregation and transformation decoupled is denoted as \textbf{InfoGraph1T}.
We fit the the empirical results to the curve of scaling law and compare the original(InfoGraph) and new implementation(InfoGraph1T) pairwise.
Figure~\ref{fig:Decouple_Loss_96} illustrates how the SSL Loss improves with more aggregations compared to the original InfoGraph, where the x-axis and y-axis indicate the number of aggregation and the SSL Loss metrics respectively. 
Our key observation is that the newly implemented \textbf{InfoGraph1T}, with decoupled aggregation and transformation, exhibits scaling behavior with scale-up number of aggregation, which is similar to the scaling behavior with scale-up number of layers exhibited in Figures~\ref{fig:model_hiv_loss},~\ref{fig:model_reddit_loss} and~\ref{fig:model_pcba_loss}.
The fitted curves of the original implementation with more transformation layers and decoupled implementation are almost overlapped.
Therefore, we can draw a conclusion that the improvement in the SSL Loss is primarily due to the model architecture or structure with more aggregations, rather than the capability with more learnable parameters for transformation.

\section{Conclusion}
In this work, we take the first step to explore the neural scaling laws on the existing Graph SSL methods.
Specifically, we try to verify the existence of two basic forms of neural scaling laws: the model scaling law and the data scaling law. 
Our attempts obtain some key observations and provide some insights for future work.
\textbf{Obs} 1 and 3 indicate that no scaling behavior can be observed in the downstream performance.
Meanwhile, \textbf{Obs} 2 and 4 indicate that scaling behavior can only be observed in the SSL loss with the increasing number of layers of the encoder in GraphSSL methods.
The above observations can draw a conclusion that the gain in the downstream performance does not correspond to SSL loss.
These results indicate that there is a huge gap between the existing SSL and downstream tasks in Graph domain.
Therefore, for further GFM design, we believe that a proper SSL task design is critical to mitigate this gap to exhibit scaling behavior on the downstream tasks.
\textbf{Obs} 5 and 6 indicate that the scaling behavior we observed is mainly caused from the characteristics of the model architecture \ie more aggregations instead of the improved capability with more learnable parameters.
These observations provide insights to future work like verifying the existence of neural scaling law on more powerful backbones \eg Graph Transformer for GraphSSL methods.
Moreover, the scaling behavior exhibited in SSL loss for contrastive methods is more consistent than generative methods.
These results suggest that SSL task design and the component design of GraphSSL methods should be considered as the key factors to reveal the potential of scaling law. 
Therefore, for further GFM design, we believe that a powerful and representative backbone is critical to be able to scale up to accommodate continuously increasing pre-training data.

Our findings shed light on the absence of the scaling behaviors of existing GraphSSL methods and point to critical components that should be considered in future design.

\bibliographystyle{unsrt}
\bibliography{0_reference}

\newpage
\appendix
\section{Dataset \label{app:dataset}}

All the datasets can be obtained from ogbg~\cite{hu2020ogb} and TUDataset~\cite{morris2020tudataset}.
Here we provide their statistics in Table~\ref{tab:data_stat}.

\begin{table}[!ht]
\centering
\caption{Datasets statistics.}
\label{tab:data_stat}
\begin{tabular}{@{}ccccc@{}}
%
Name           & \# Graphs & \# Avg. nodes & \# Avg. edges & Metric   \\ \midrule
ogbg-hiv       & 41,127   & 25.5         & 27.5         & ROC-AUC  \\
ogbg-molpcba   & 437,929  & 26.0         & 28.1         & AP       \\
ogbg-ppa       & 158,100  & 243.4        & 2,266.1      & Accuracy \\
Reddit-Threads & 203,088  & 23.93        & 24.99        & Accuracy \\
ZINC-Full      & 249,456  & 23.15        & 24.90        & RMSE     \\ \bottomrule

\end{tabular}
\end{table}

\section{Experiment Specific Settings \label{app:settings}}
All hyper-parameter configuration files for the methods used in this study are provided in the code repository. Below, we outline the settings for some general hyper-parameters:

\begin{itemize}
    \item \textbf{Hidden Size and Number of Layers}: For all methods involved in data scaling experiments, the hidden size is set to 32, and the number of layers is set to 2.
    \item \textbf{Learning Rate and Optimizer}: Across all experiments, the initial learning rate is set to 0.001, and the Adam optimizer~\cite{kingma2014adam} is employed for training.
    \item \textbf{Graph Classification Task}: For downstream evaluation, an SVM classifier is trained, with the C parameter selected via grid search over the range [0.001, 0.01, 0.1, 1, 10, 100, 1000].
    \item \textbf{Graph Regression Task}: For downstream evaluation, a two-layer MLP is trained with a hidden size matching that of the pre-trained encoder.
\end{itemize}

\section{More Results on Data Scaling\label{app:data_scaling}}

\subsection{Data Scaling on Downstream Performance}
In this section, we are providing additional results on the deferred ogbg-molpcba for graph classification task,
and ZINC-Full dataset for graph regression task.

Our observations from these results remain the same as that we illustrated in the main content that there is no scaling behavior can be observed on downstream performance with the gradually scaled-up pretraining data.

\begin{figure}[htbp]
    \centering
    \subfigure[GraphCL $R^2$=0.0]{
        \includegraphics[width=0.23\textwidth]{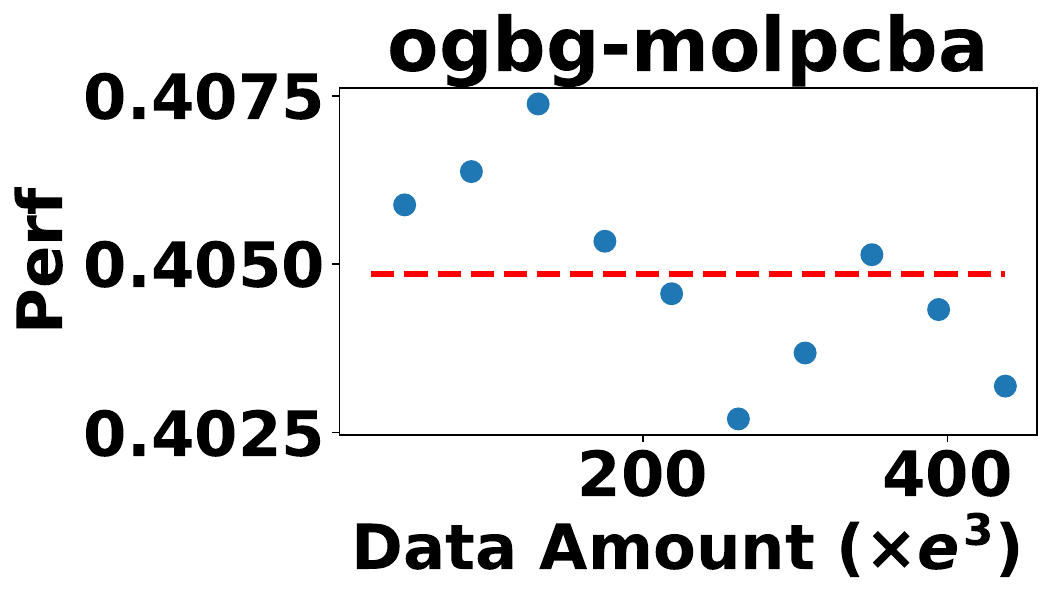} 
        \label{fig:data_GraphCL_ogbg-molpcba_perf}
    }
    \subfigure[GraphMAE $R^2$=0.0]{
        \includegraphics[width=0.23\textwidth]{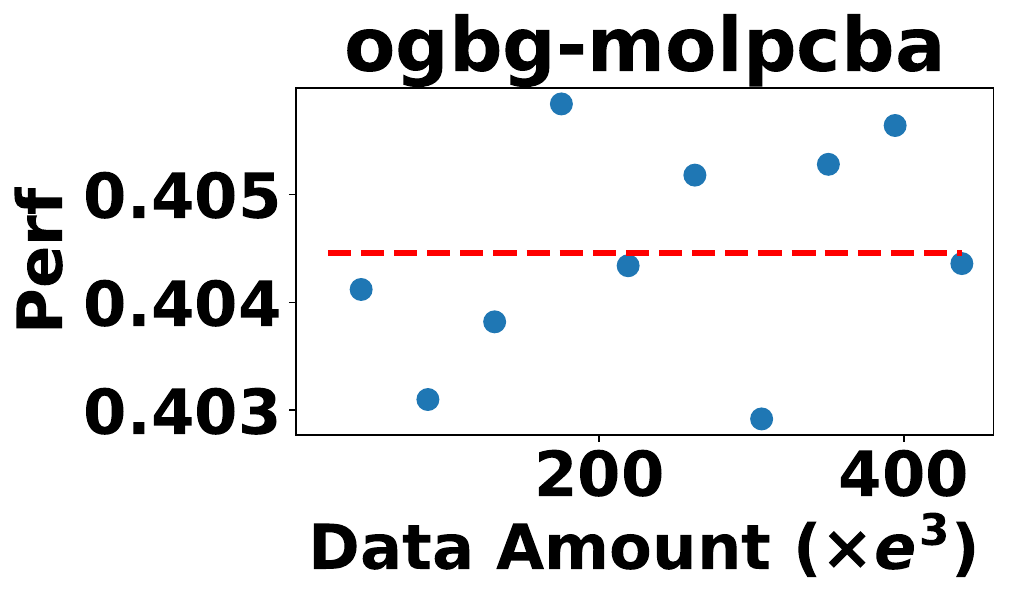}
        \label{fig:data_GraphMAE_ogbg-molpcba_perf}
    }
    \subfigure[InfoGraph $R^2$=0.0]{
        \includegraphics[width=0.23\textwidth]{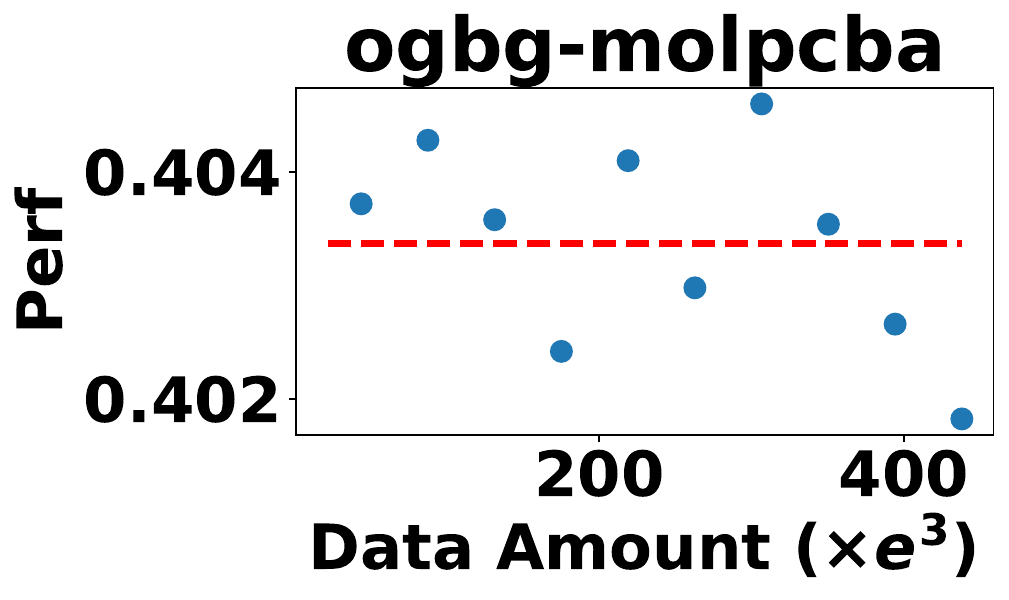} 
        \label{fig:data_InfoGraph_ogbg-molpcba_perf}
    }
    \subfigure[JOAO $R^2$=0.03]{
        \includegraphics[width=0.23\textwidth]{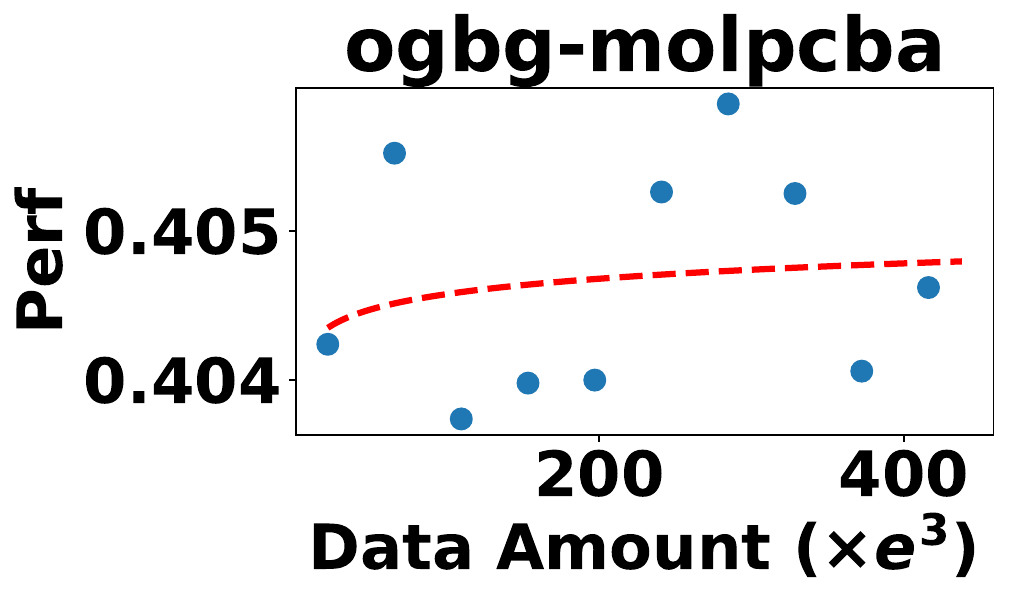}
        \label{fig:data_JOAO_oogbg-molpcba_perf}
    }
    \caption{Data Scaling of Performance on ogbg-molpcba. x-axis indicates the data amount used for pre-training and y-axis indicates the downstream performance. No obvious scaling behavior can be observed.}
    \label{fig:Data_Perf_pcba}
\end{figure}

\begin{figure}[htbp]
    \centering
    \subfigure[GraphCL $R^2$=0.11]{
        \includegraphics[width=0.23\textwidth]{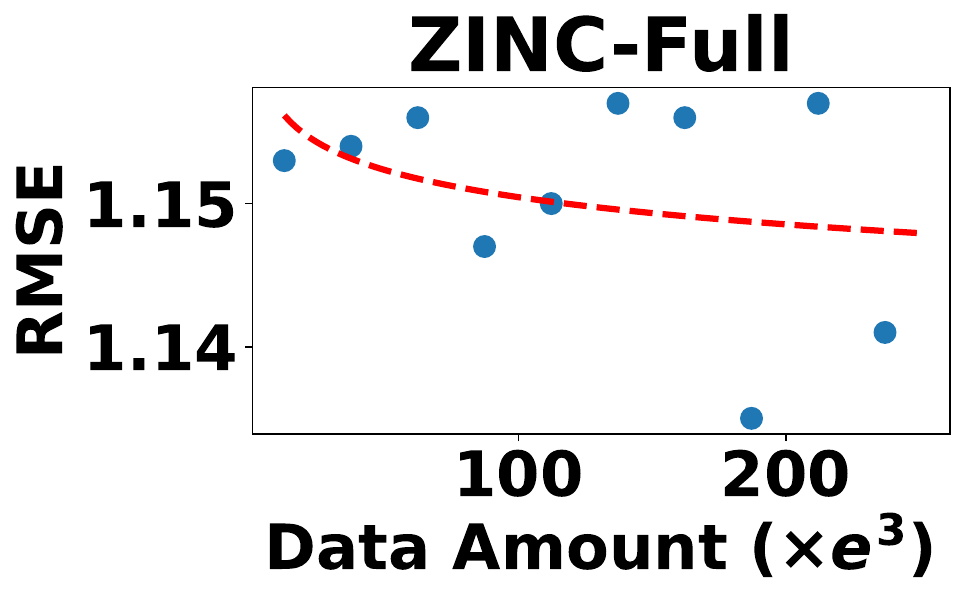} 
        \label{fig:data_GraphCL_ZINC_perf}
    }
    \subfigure[GraphMAE $R^2$=0.0]{
        \includegraphics[width=0.23\textwidth]{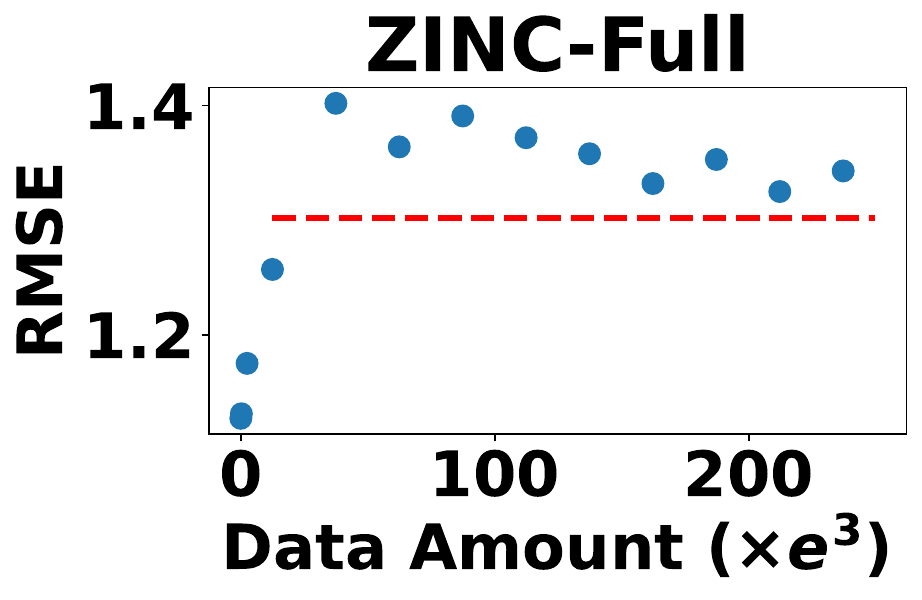}
        \label{fig:data_GraphMAE_ZINC_perf}
    }
    \subfigure[InfoGraph $R^2$=0.80]{
        \includegraphics[width=0.23\textwidth]{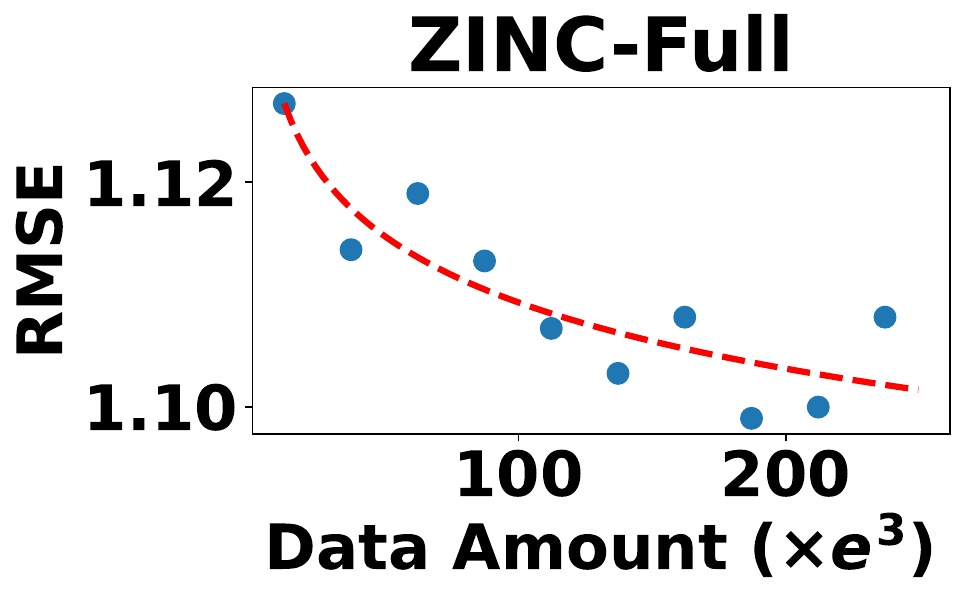} 
        \label{fig:data_InfoGraph_ZINC_perf}
    }
    \subfigure[JOAO $R^2$=0.39]{
        \includegraphics[width=0.23\textwidth]{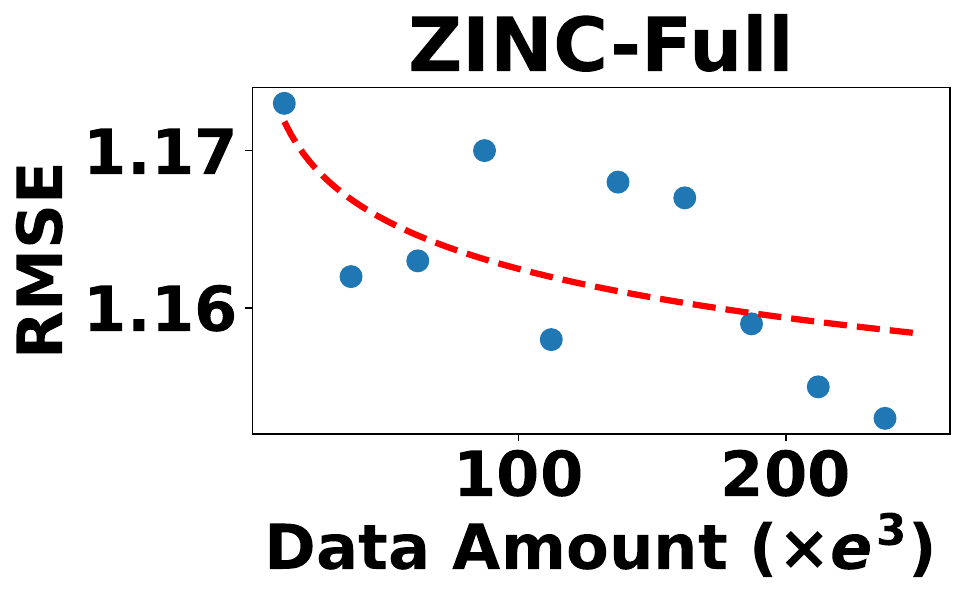}
        \label{fig:data_JOAO_ZINC_perf}
    }
    \caption{Data Scaling of Performance on ZINC-Full. x-axis indicates the data amount used for pre-training and y-axis indicates the downstream performance. No obvious scaling behavior can be observed.}
    \label{fig:Data_Perf_ZINC}
\end{figure}

\subsection{Data Scaling on SSL Loss}

In this section, we are providing additional results of Data Scaling on SSL Loss on the deferred ogbg-molpcba and ZINC-Full dataset for graph regression task.

Our observations from these results remain the same as that we illustrated in the main content that scaling behavior can be observed on downstream SSL loss with the gradually scaled-up pre-training data.

\begin{figure}[!htbp]
    \centering
    \subfigure[InfoGraph $R^2=0.93$]{
        \includegraphics[width=0.23\textwidth]{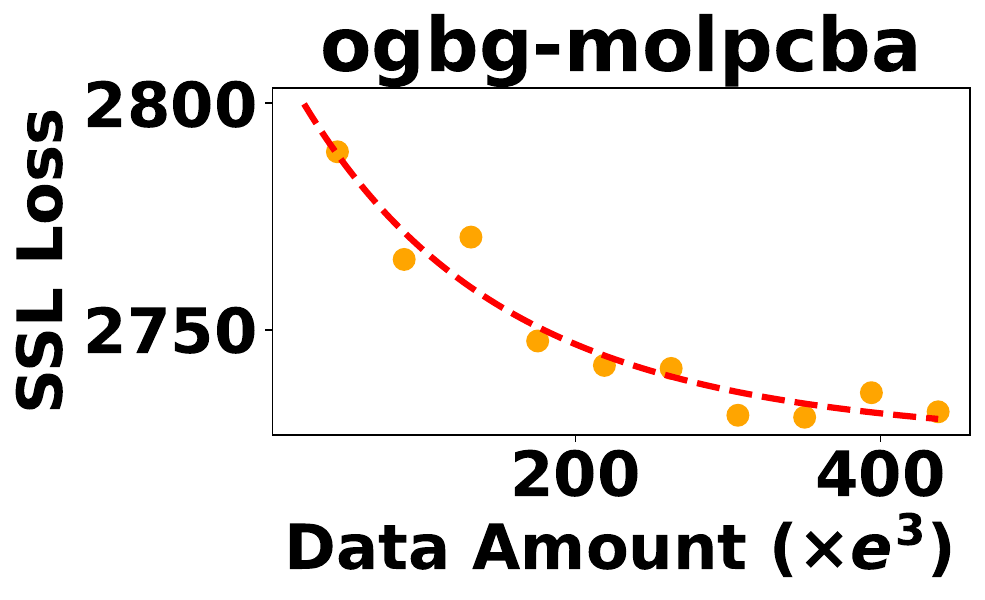}
        \label{fig:data_InfoGraph_ogbg-molpcba_loss}
    }
    \subfigure[GraphCL $R^2=0.58$]{
        \includegraphics[width=0.23\textwidth]{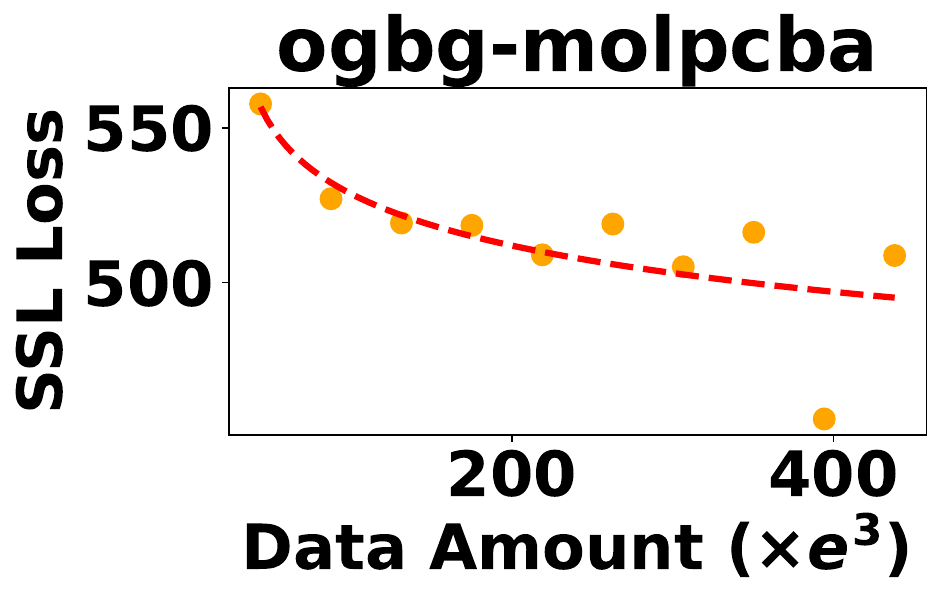}
        \label{fig:data_GraphCL_ogbg-molpcba_loss}
    }
    \subfigure[JOAO $R^2=0.01$]{
        \includegraphics[width=0.23\textwidth]{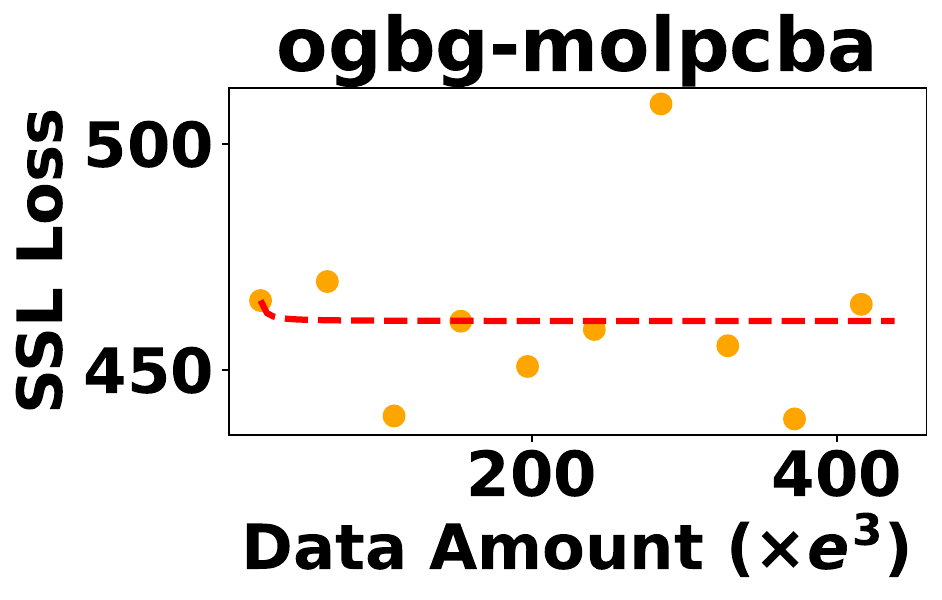}
        \label{fig:data_JOAO_ogbg-molpcba_loss}
    }
    \subfigure[GraphMAE $R^2=0.0$]{
        \includegraphics[width=0.23\textwidth]{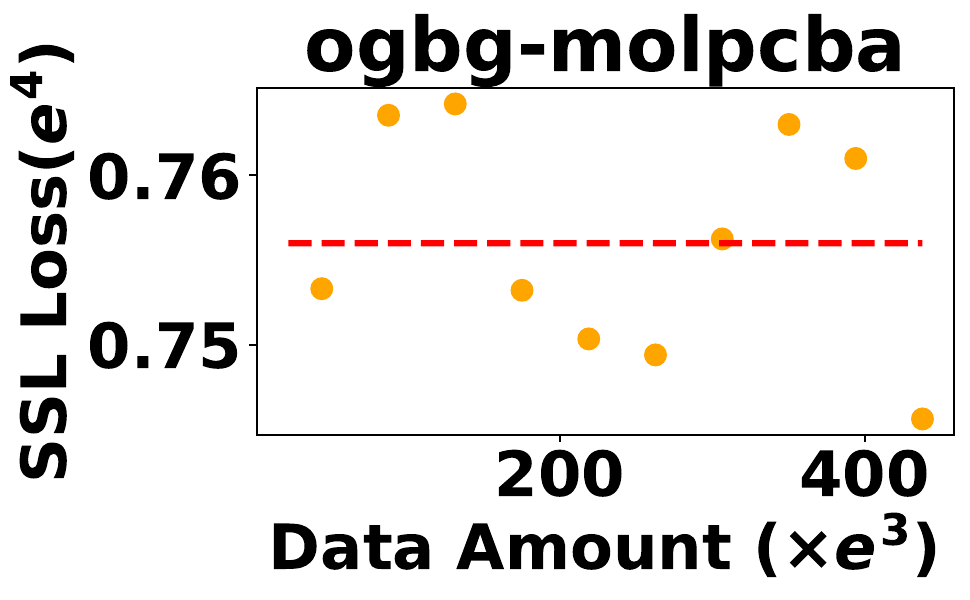}
        \label{fig:data_GraphMAE_ogbg-molpcba_loss}
    }
    \caption{Data Scaling of SSL Loss on ogbg-molpcba.x-axis indicates the data amount used for pre-training and y-axis indicates the SSL Loss on the held-out test data. More obvious scaling behavior can be observed on InfoGraph and GraphCL compared to performance.}
    \label{fig:data-loss-pcba}
\end{figure}

\begin{figure}[!htbp]
    \centering
    \subfigure[InfoGraph $R^2=0.99$]{
        \includegraphics[width=0.23\textwidth]{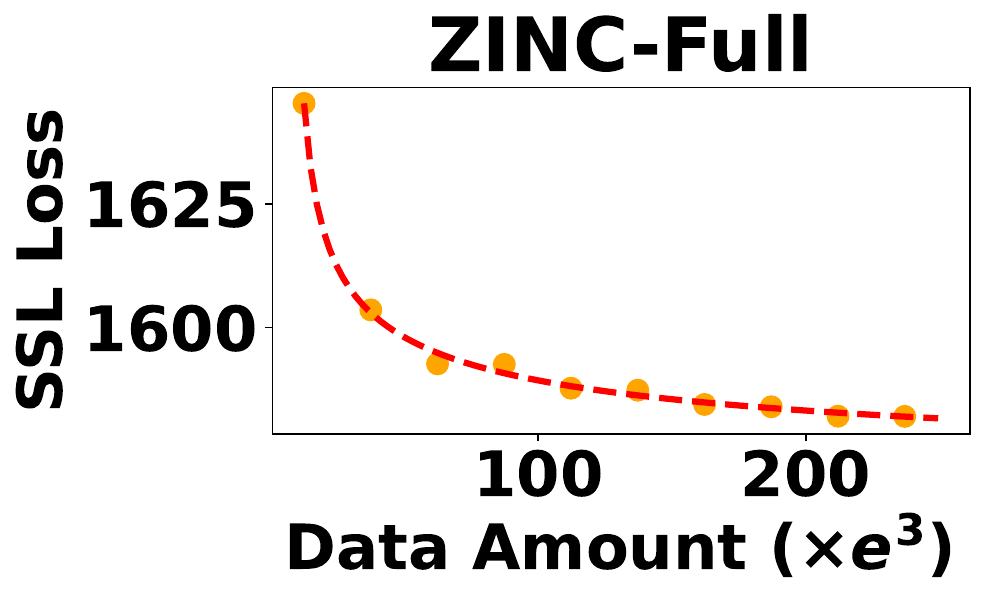}
        \label{fig:data_InfoGraph_ZINC_loss}
    }
    \subfigure[GraphCL $R^2=0.98$]{
        \includegraphics[width=0.23\textwidth]{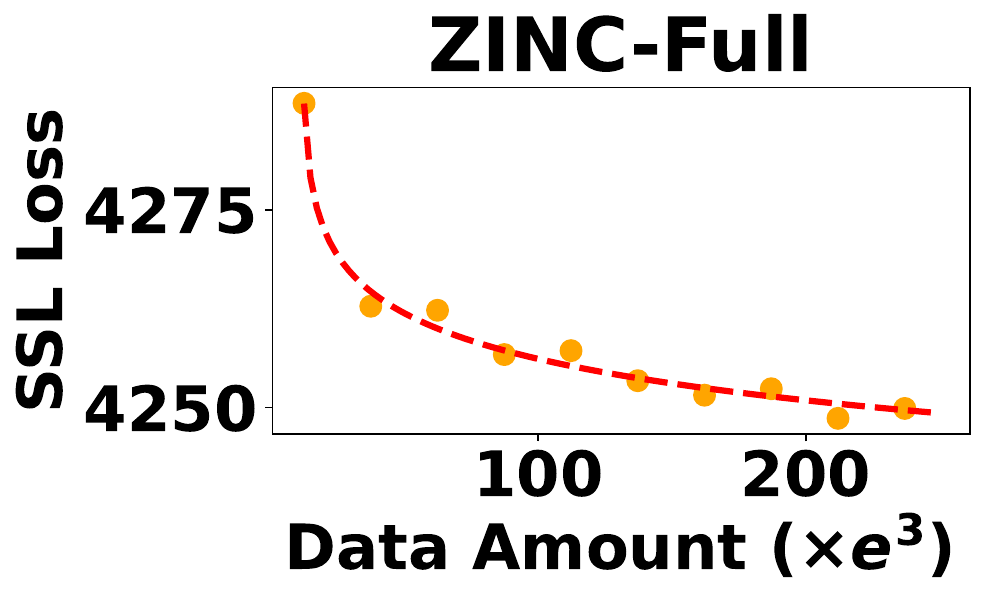}
        \label{fig:data_GraphCL_ZINC_loss}
    }
    \subfigure[JOAO $R^2=0.98$]{
        \includegraphics[width=0.23\textwidth]{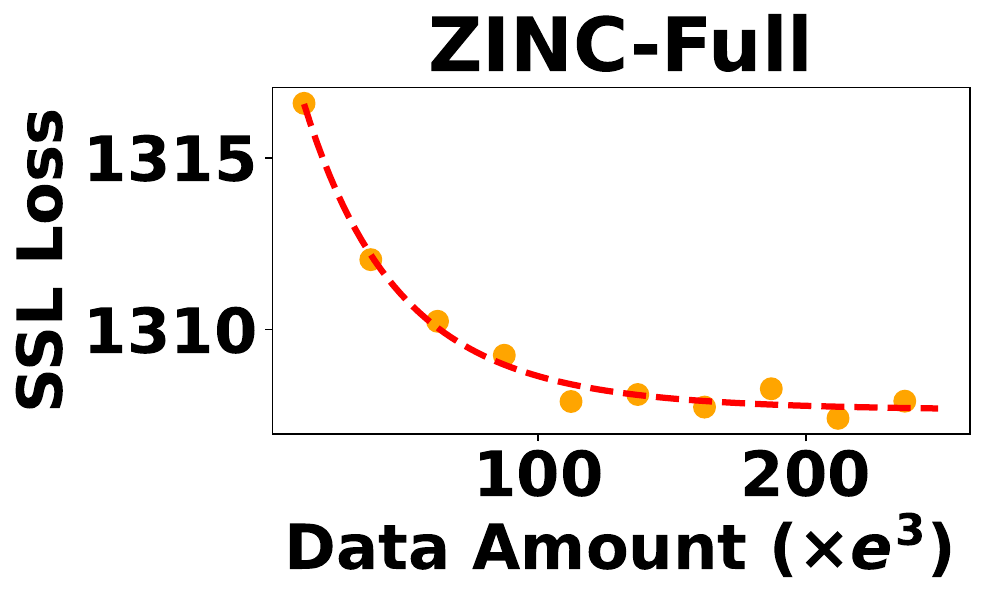}
        \label{fig:data_JOAO_ZINC_loss}
    }
    \subfigure[GraphMAE $R^2=0.97$]{
        \includegraphics[width=0.23\textwidth]{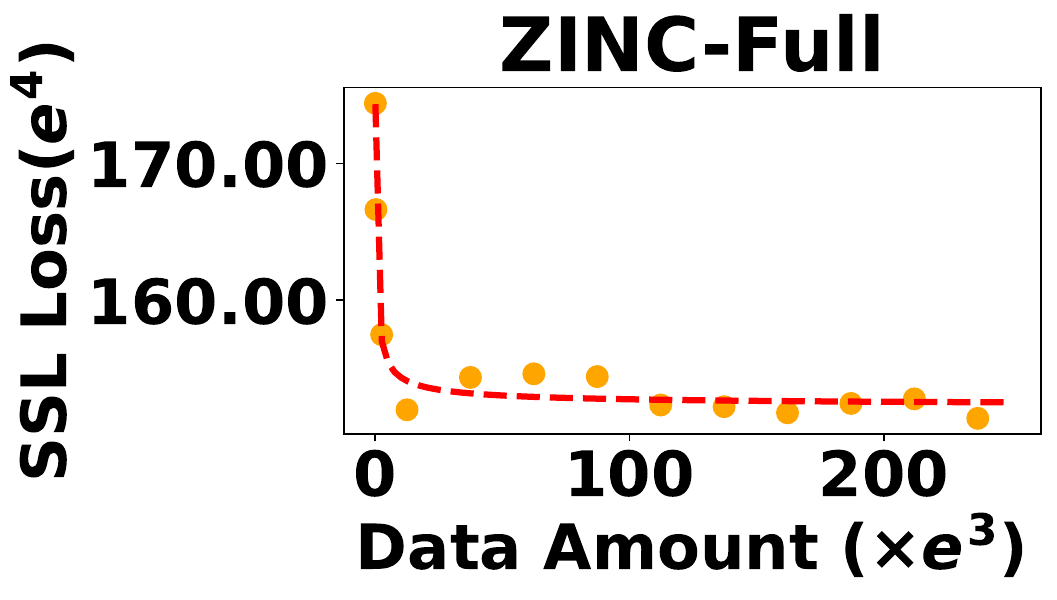}
        \label{fig:data_GraphMAE_ZINC_loss}
    }
    \caption{Data Scaling of SSL Loss on ZINC-Full.x-axis indicates the data amount used for pre-training and y-axis indicates the SSL Loss on the held-out test data. More obvious scaling behavior can be observed compared to performance.}
    \label{fig:data-loss-ZINC}
\end{figure}

\section{More Results on Model Scaling\label{app:model_scaling}}

\subsection{Model scaling on Downstream performance}

In this section, we are providing additional results on the ogbg-molpcba, ogbg-ppa and ZINC-Full datasets respectively.
ZINC-Full is evaluated by graph regression tasks while the others are evaluated by graph classification task.

Our observations from these results remain the same as that we illustrated in the main content that there is no obvious scaling behavior can be observed under model scaling via increasing number of layers or the hidden size of the model.

\begin{figure}[!htbp]
    \centering
    \includegraphics[width=\textwidth]{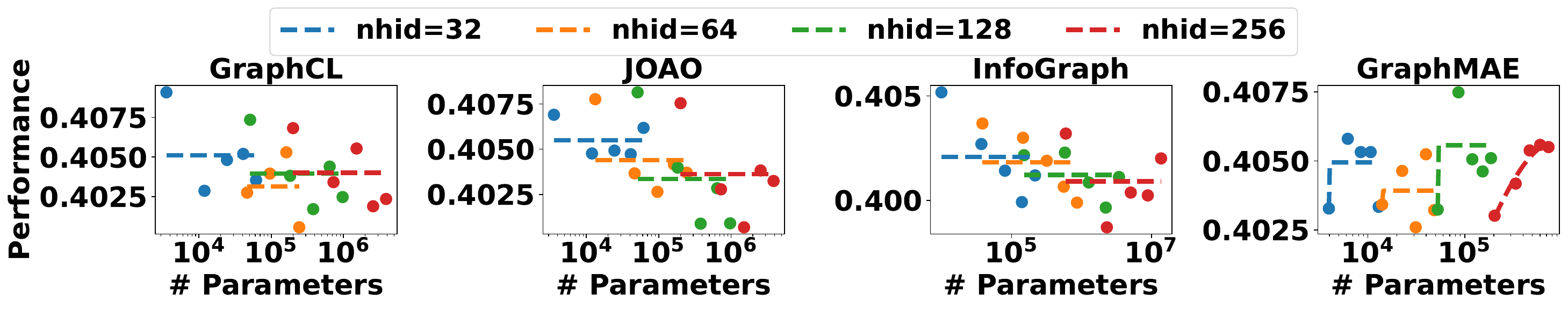}
    \caption{\small{Performance on ogbg-molpcba. x-axis denotes the total number of parameters and y-axis denotes the downstream performance. No obvious scaling behaviour can be observed on all methods.The $R^2$ values for each method are listed as follows.GraphCL:0.0,JOAO:0.0,InfoGraph:0.0,GraphMAE:0.46}}
    \label{fig:model_pcba_perf}
\end{figure}

\begin{figure}[!htbp]
    \centering
    \includegraphics[width=\textwidth]{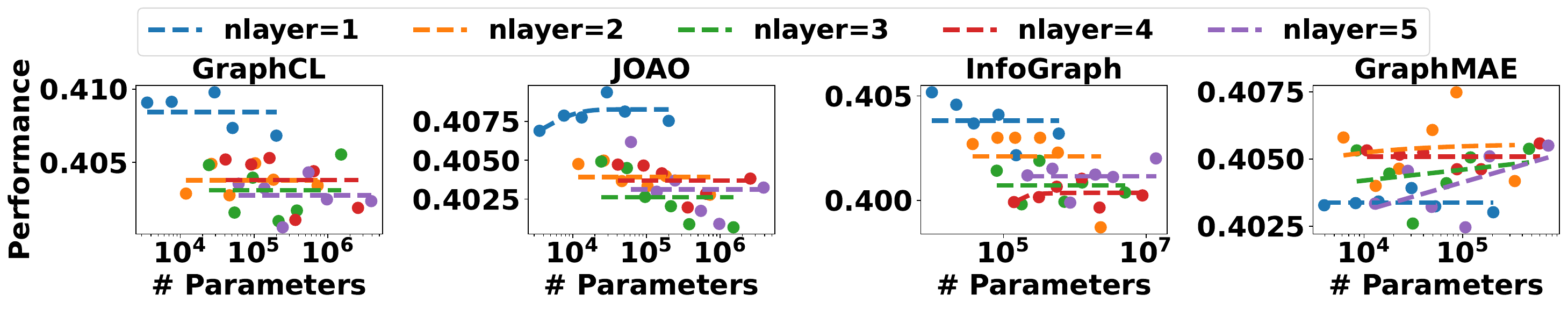}
    \caption{\small{Performance on ogbg-molpcba grouped by layers. x-axis denotes the total number of parameters and y-axis denotes the downstream performance. No obvious scaling behaviour can be observed.The $R^2$ values for each method are listed as follows,GraphCL:0.10,JOAO:0.09,InfoGraph:0.02,GraphMAE:0.18}}
    \label{fig:model_pcba_perf_layer}
\end{figure}

\begin{figure}[!htbp]
    \centering
    \includegraphics[width=\textwidth]{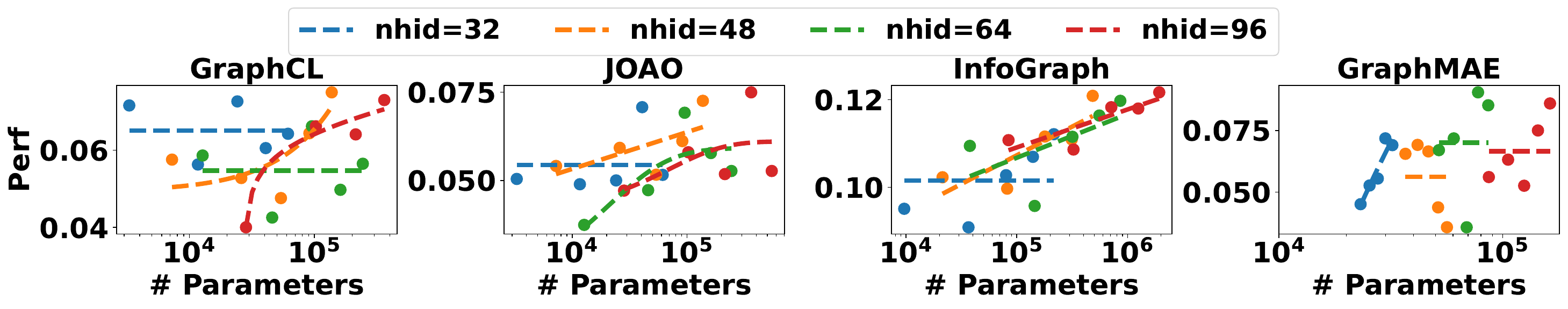}
    \caption{\small{Performance on ogbg-ppa. x-axis denotes the total number of parameters and y-axis denotes the downstream performance. No obvious scaling behaviour can be observed on all methods.The $R^2$ values for each method are listed as follows.GraphCL:0.39,JOAO:0.33,InfoGraph:0.43,GraphMAE:0.22}}
    \label{fig:model_ppa_perf}
\end{figure}

\begin{figure}[!htbp]
    \centering
    \includegraphics[width=\textwidth]{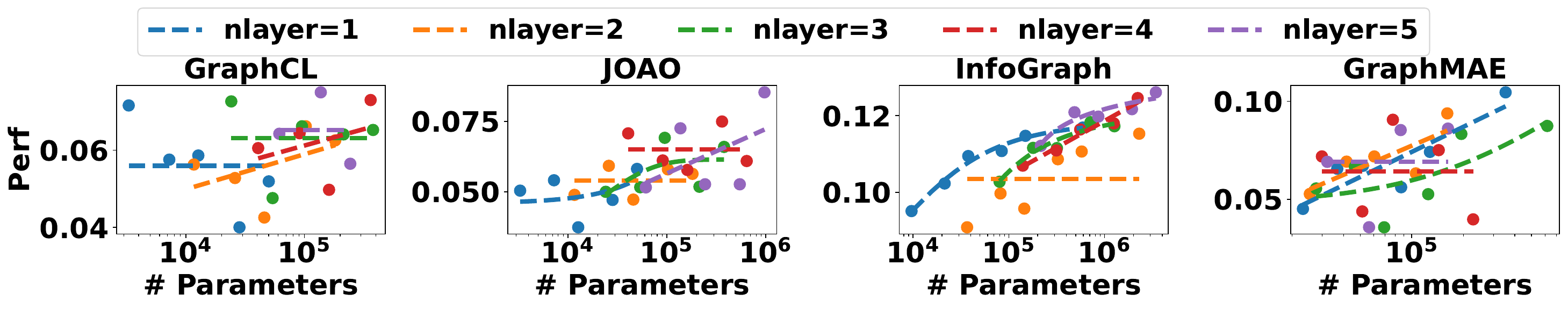}
    \caption{\small{Performance on ogbg-ppa grouped by layers. x-axis denotes the total number of parameters and y-axis denotes the downstream performance. No obvious scaling behaviour can be observed.The $R^2$ values for each method are listed as follows,GraphCL:0.06,JOAO:0.14,InfoGraph:0.75,GraphMAE:0.38}}
    \label{fig:model_ppa_perf_layer}
\end{figure}

\begin{figure}[!htbp]
    \centering
    \includegraphics[width=\textwidth]{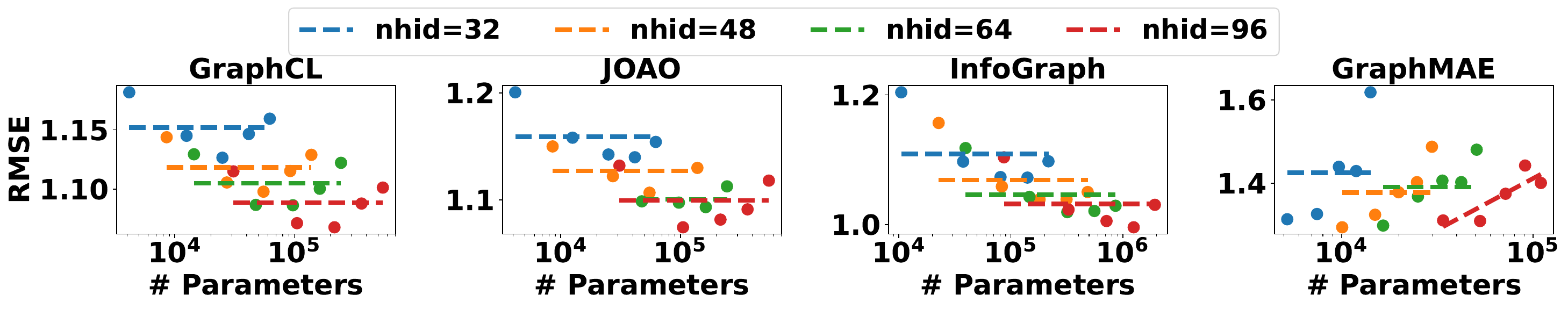}
    \caption{\small{Performance on ZINC-Full. x-axis denotes the total number of parameters and y-axis denotes the downstream performance. No obvious scaling behaviour can be observed on all methods.The $R^2$ values for each method are listed as follows.GraphCL:0.0,JOAO:0.0,InfoGraph:0.0,GraphMAE:0.46}}
    \label{fig:model_ZINC_perf}
\end{figure}

\begin{figure}[!htbp]
    \centering
    \includegraphics[width=\textwidth]{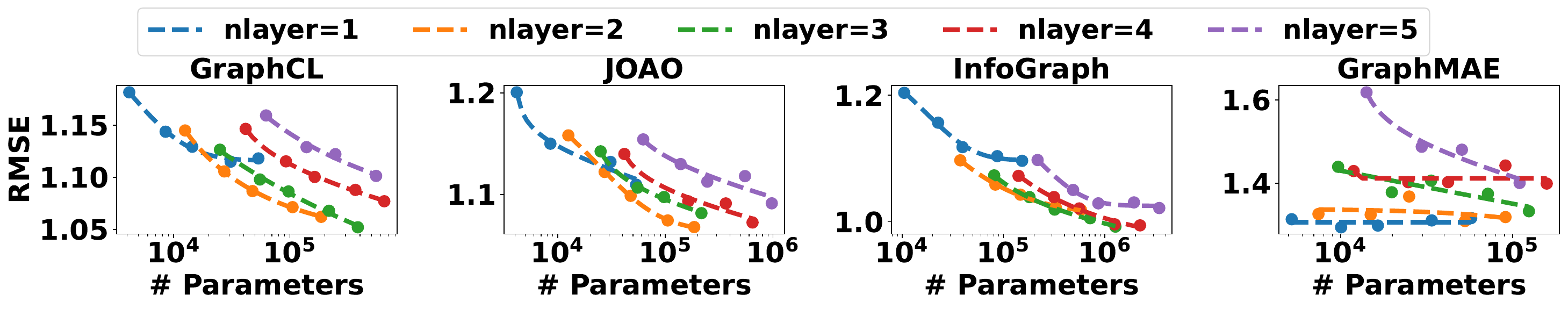}
    \caption{\small{Performance on ZINC-Full grouped by layers. x-axis denotes the total number of parameters and y-axis denotes the downstream performance. No obvious scaling behaviour can be observed.The $R^2$ values for each method are listed as follows,GraphCL:0.99,JOAO:0.96,InfoGraph:0.99,GraphMAE:0.41. Kindly note that $R^2$ does not indicate the existence of scaling effect instead of the fitting quality.}}
    \label{fig:model_ZINC_perf_layer}
\end{figure}

\subsection{Model scaling on SSL loss}

In this section, we are providing additional results of the Model scaling on SSL loss on the ogbg-molpcba, ogbg-ppa and ZINC-Full datasets respectively.

Our observations from these results remain the same as that we illustrated in the main content that there is obvious scaling behavior on downstream SSL loss that can only be observed under model scaling via increasing number of layers of the model.

\begin{figure}[!htbp]
    \centering
    \includegraphics[width=0.95\textwidth]{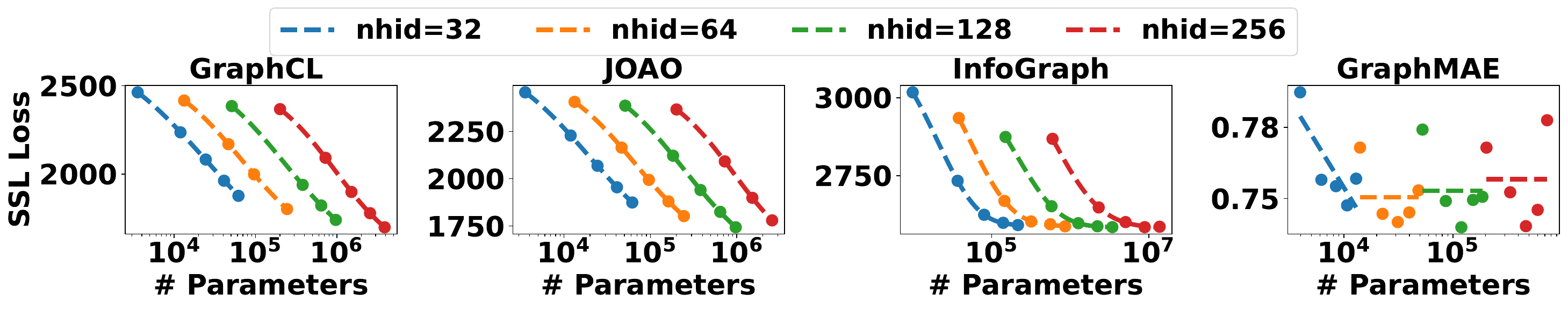}
    \caption{SSL Loss on ogbg-molpcba.Obvious scaling behaviour can be observed except GraphMAE.  x-axis denotes the total number of parameters and y-axis denotes SSL Loss on the held-out test set. The $R^2$ values for each method are listed as follows. GraphCL:0.99, JOAO:0.99, InfoGraph:0.99, GraphMAE:0.17}
    \label{fig:model_pcba_loss}
\end{figure}

\begin{figure}[!htbp]
    \centering
    \includegraphics[width=0.95\textwidth]{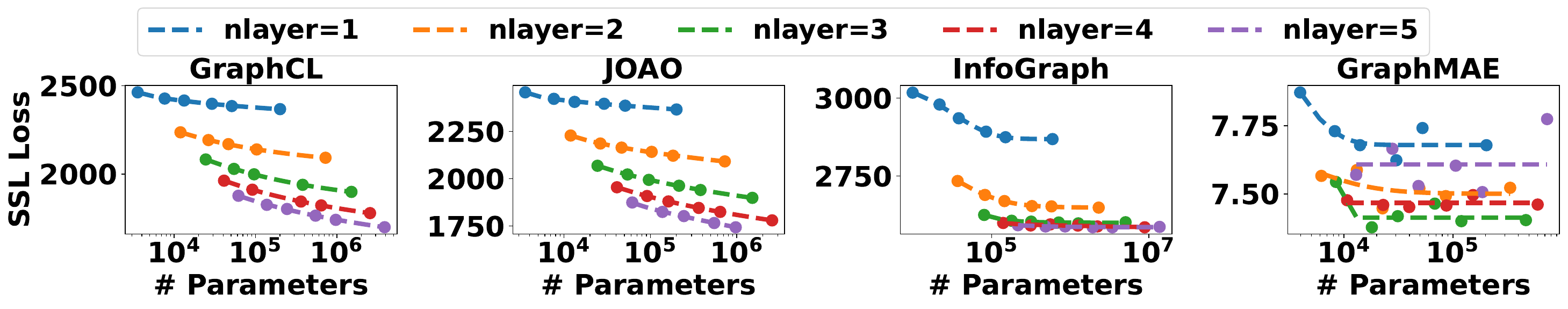}
    \caption{SSL Loss on ogbg-molpcba grouped by layer. No obvious scaling behaviour can be observed. x-axis denotes the total number of parameters and y-axis denotes SSL Loss on the held-out test set. The $R^2$ values for each method are listed as follows. GraphCL:0.99, JOAO:0.99, InfoGraph:0.95, GraphMAE:0.38}
    \label{fig:model_pcba_loss_layer}
\end{figure}

\begin{figure}[!htbp]
    \centering
    \includegraphics[width=0.95\textwidth]{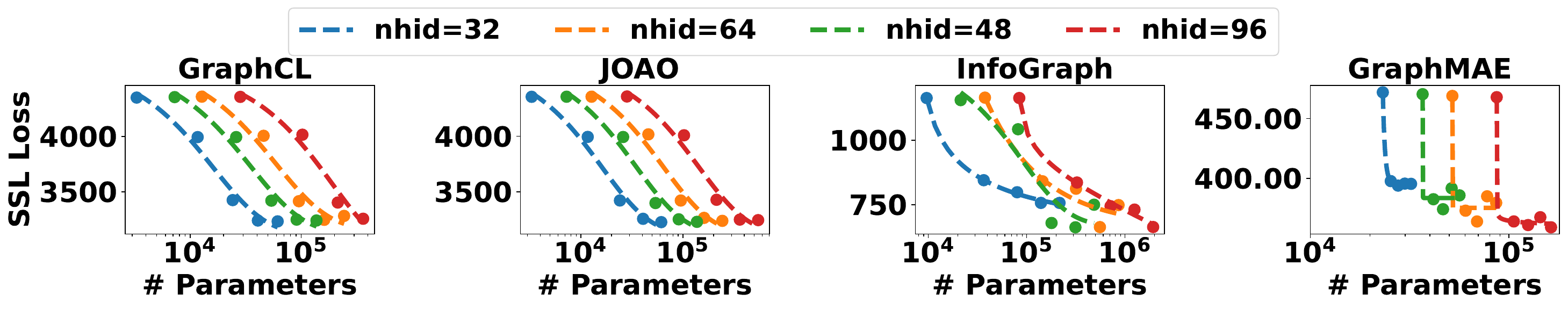}
    \caption{SSL Loss on ogbg-ppa.Obvious scaling behaviour can be observed.  x-axis denotes the total number of parameters and y-axis denotes SSL Loss on the held-out test set. The $R^2$ values for each method are listed as follows. GraphCL:0.97, JOAO:0.97, InfoGraph:0.95, GraphMAE:0.98}
    \label{fig:model_ppa_loss}
\end{figure}

\begin{figure}[!htbp]
    \centering
    \includegraphics[width=0.95\textwidth]{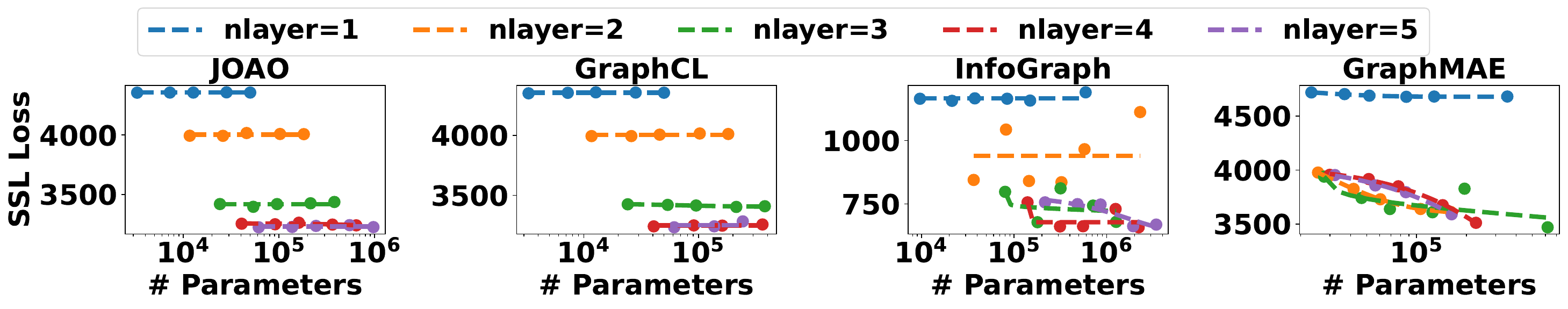}
    \caption{SSL Loss on ogbg-ppa grouped by layer. No obvious scaling behaviour can be observed. x-axis denotes the total number of parameters and y-axis denotes SSL Loss on the held-out test set. The $R^2$ values for each method are listed as follows. GraphCL:0.18, JOAO:0.11, InfoGraph:0.34, GraphMAE:0.92}
    \label{fig:model_ppa_loss_layer}
\end{figure}

\begin{figure}[!htbp]
    \centering
    \includegraphics[width=0.95\textwidth]{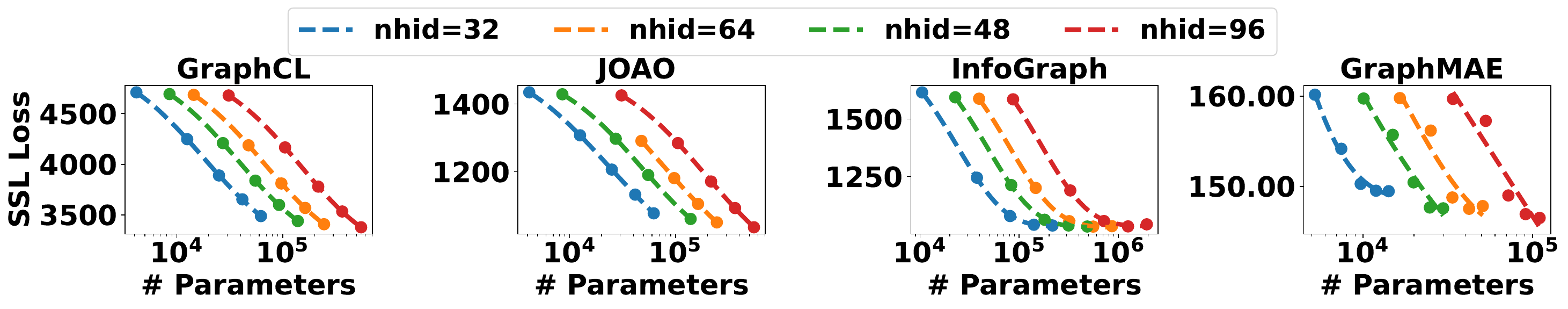}
    \caption{SSL Loss on ZINC-Full.Obvious scaling behaviour can be observed on all methods.  x-axis denotes the total number of parameters and y-axis denotes SSL Loss on the held-out test set. The $R^2$ values for each method are listed as follows. GraphCL:0.99, JOAO:0.99, InfoGraph:0.99, GraphMAE:0.95}
    \label{fig:model_ZINC_loss}
\end{figure}

\begin{figure}[!htbp]
    \centering
    \includegraphics[width=0.95\textwidth]{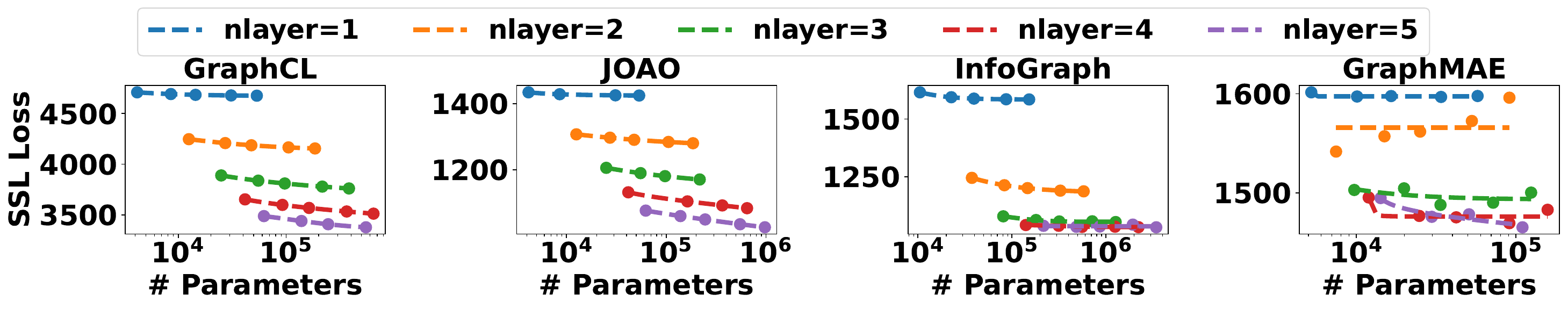}
    \caption{SSL Loss on ZINC-Full grouped by layer. No obvious scaling behaviour can be observed. x-axis denotes the total number of parameters and y-axis denotes SSL Loss on the held-out test set. The $R^2$ values for each method are listed as follows. GraphCL:0.99, JOAO:0.99, InfoGraph:0.78, GraphMAE:0.58}
    \label{fig:model_ZINC_loss_layer}
\end{figure}

\section{Deferred details about GraphCL/JOAO SSL method}\label{sec:model-GraphCL/JOAO}
Compared with InfoGraph, there are two major differences between GraphCL/JOAO and InfoGraph.
(1) The SSL Loss (2) The strategy for constructing the augmented view for contrastive learning.

We first investigate the influence of loss.
InfoGraph is using JSD Loss while GraphCL and JOAO are using InfoNCE loss.
As this is the most obvious difference between the methods, we switch the SSL Loss, which can be considered as switching a single component between two different frameworks.
The initial observation indicates that the instability of GraphCL and JOAO remained the same for the revised version while the revised InfoGraph was still stable. Consequently, our key conclusion from the above results is that the SSL Loss is not the factor that affects the stability.

We further investigated the difference in generating the augmented views. By fixing the randomness in utilizing augmenters to generate augmented views for contrastive learning. The rest of the settings are the same as the model scaling settings with fixed hidden size.
After fixing the randomness in augmented view generation and selecting a proper contrastive strategy, GraphCL and JOAO obtain more stable results, where more obvious scaling behavior can be exhibited.
Meanwhile, their downstream performances are still almost overlapped with the ones with randomly selected data augmentations.
These results also support our conclusions that the gap between SSL and downstream tasks blocks the SSL methods from improving on downstream performance corresponding to SSL loss and the component design is critical for exhibiting scaling behavior for the future Graph Foundation Model design.

\end{document}